\documentclass{article}

 \usepackage[final,nonatbib]{neurips_2019}

\usepackage[utf8]{inputenc} 
\usepackage[T1]{fontenc}    
\usepackage{url}            
\usepackage{booktabs}       
\usepackage{amsfonts}       
\usepackage{nicefrac}       
\usepackage{microtype}      
\usepackage{graphicx,color}
\usepackage{mathrsfs}
\usepackage{amsmath}
\usepackage{amsthm}
\usepackage{amssymb,amsfonts}
\usepackage{cite}
\usepackage{algorithm}
\usepackage[noend]{algpseudocode}
\usepackage{textcomp}
\usepackage{subfigure}
\usepackage{dsfont}
\usepackage{bbm}
\usepackage{array}
\usepackage[table]{xcolor} 
\usepackage{yfonts}
\usepackage{enumitem}
\usepackage{appendix}

\usepackage{xr-hyper}
\usepackage{hyperref}

\newcommand\aamsout{\bgroup\markoverwith{\textcolor{violet}{\rule[0.5ex]{2pt}{1pt}}}\ULon}

\newcommand{\real}{\mathbb{R}}
\newcommand{\realnonnegative}{\mathbb{R}_{\geq 0}}

\newcommand{\transpose}{\mathsf{T}} 

\newcommand{\mc}{\mathcal}

\DeclareSymbolFont{bbold}{U}{bbold}{m}{n}
\DeclareSymbolFontAlphabet{\mathbbold}{bbold}

\newcommand{\lip}{\mathrm{lip}}

\newcommand{\Lip}{\mathrm{Lip}}

\newcommand{\ess}{\mathrm{ess}}

\DeclareMathOperator{\id}{Id}

\newtheorem{theorem}{Theorem}[section]

\title{Lipschitz Bounds and Provably Robust Training by Laplacian Smoothing}

\author{%
 Vishaal~Krishnan \\
  Mechanical Engineering Department \\
  University of California Riverside\\
  \texttt{vishaalk@ucr.edu} \\
  \And
  Abed AlRahman~Al Makdah\\
  Electrical \& Computer Engineering Department\\
  University of California Riverside\\
  \texttt{aalmakdah@engr.ucr.edu} \\
  \And
  Fabio~Pasqualetti \\
  Mechanical Engineering Department\\
  University of California Riverside\\
  \texttt{fabiopas@engr.ucr.edu} \\
}
\graphicspath{../figs/}
\begin{document}

\maketitle

\begin{abstract}  
  In this work we propose a graph-based learning framework to train
  models with provable robustness to adversarial perturbations. In
  contrast to regularization-based approaches, we formulate the
  adversarially robust learning problem as one of loss minimization
  with a Lipschitz constraint, and show that the saddle point of the
  associated Lagrangian is characterized by a Poisson equation with
  weighted Laplace operator. Further, the weighting for the Laplace
  operator is given by the Lagrange multiplier for the Lipschitz
  constraint, which modulates the sensitivity of the minimizer to
  perturbations. We then design a provably robust training scheme
  using graph-based discretization of the input space and a
  primal-dual algorithm to converge to the Lagrangian's saddle
  point. Our analysis establishes a novel connection between elliptic
  operators with constraint-enforced weighting and adversarial
  learning.
  We also study the complementary problem of improving the robustness
  of minimizers with a margin on their loss, formulated as a
  loss-constrained minimization problem of the Lipschitz constant. We
  propose a technique to obtain robustified minimizers, and evaluate
  fundamental Lipschitz lower bounds by approaching Lipschitz constant
  minimization via a sequence of gradient $p$-norm minimization
  problems.
  Ultimately, our results show that, for a desired nominal
  performance, there exists a fundamental lower bound on the
  sensitivity to adversarial perturbations that depends only on the
  loss function and the data distribution, and that improvements in
  robustness beyond this bound can only be made at the expense of
  nominal performance.  Our training schemes provably achieve these
  bounds both under constraints on performance and~robustness.
\end{abstract}

\section{Introduction}\label{sec: introduction}
Sensitivity to adversarial perturbations is one of the main
limitations of data-driven models, and a hurdle to their deployment in
safety-critical applications.
Improving adversarial robustness requires adjusting the worst-case
sensitivity of the data-driven input-output map, which is
characterized by its Lipschitz constant.
Training under a Lipschitz regularization or constraint is therefore a
natural way of improving adversarial robustness, which has led to many
works on the subject~\cite{HG-EF-BP-MC:18, CF-JC-BA-AO:18}.
Yet, a fundamental understanding of the limitations of this 
approach, as well as a general framework for training models that are
provably robust to adversarial perturbations, remain critically
lacking.

Motivated by this need, we consider the problem of adversarially robust learning, 
formulated as a loss minimization problem with a Lipschitz constraint:
\begin{align}\label{eq:adv_robust_expected_loss_min}
  \inf_{f \in \Lip(\mathbb{X} ; \mathbb{Y}) }~ 
  \underbrace{\mathbb{E}_{(x,y) \sim \sigma} \left[ \ell
  \left( f(x), y \right) \right]}_{\triangleq L_{\sigma}(f)},
  \qquad  \text{s.t.}~~\lip(f) \leq \alpha  ,
\end{align}
where $\mathbb{X}$ and~$\mathbb{Y}$ are the
input and output spaces equipped with distance functions,
$\ell$ is the loss function for the learning problem,
$\sigma$ the data-generating distribution and
the search space  
is the space~$\Lip(\mathbb{X}; \mathbb{Y})$ of Lipschitz-continuous 
maps from~$\mathbb{X}$ to~$\mathbb{Y}$ 
with an upper bound~$\alpha$
on the Lipschitz constant.
This class of problems includes, for instance, the problem of image
classification with a constraint on the Lipschitz constant of the
classifier.
In this case,~$x$ denotes an image,~$y$ a probability vector over the
space of labels and $\sigma$ captures the relation between images and
labels.
In \eqref{eq:adv_robust_expected_loss_min}, we do not restrict our
attention to any finite-dimensional subspace
of~$\Lip(\mathbb{X}; \mathbb{Y})$, as done when a particular machine
learning model is chosen (for instance, neural network, where the
dimension of the search space is specified by the network structure).
Instead, we focus on the infinite-dimensional learning problem to
derive insights and fundamental bounds for the underlying adversarial
learning problem.
Finally, imposing a hard constraint on the Lipschitz constant (as
opposed to a regularization term) allows us to provide hard guarantees
on the robustness of the minimizer to adversarial perturbations.

\textbf{Contributions.} In this paper we characterize fundamental
robustness bounds for machine learning algorithms, and design provably
robust training schemes. Our approach creates, to the best of our
knowledge, a novel and useful bridge between the nascent theory of
provably robust learning and the classic theories of elliptic
operators, partial differential equations, and numerical integration.
The technical contributions of this paper are twofold. First, in
Section~\ref{sec: robust_min} we consider
Problem~\eqref{eq:adv_robust_expected_loss_min} of designing a
data-driven map to minimize the loss function, with a desired bound on
the map's Lipschitz constant.  Under assumptions on strict convexity
of the loss function and compactness of the input and output spaces,
we show that the problem has a unique minimizer and characterize the
saddle point of the corresponding Lagrangian for the problem as the
(weak) solution to a Poisson partial differential equation involving a
weighted Laplace operator, with the weighting given by the Lagrange
multiplier for the constraint.  This result provides key insights into
the nature of the optimal data-driven map satisfying robustness
constraints.  We then design a provably robust training scheme based
on a graph discretization of the domain to numerically solve for the
minimizer of the problem.

Second, we consider the problem of minimizing the Lipschitz constant
of a data-driven map with a guaranteed bound (margin) on its loss.  We
show that the Lipschitz constant is tightly and inversely related to
the loss, thereby revealing a fundamental tradeoff between the
robustness of a data-driven map and its performance.  This result
implies that the Lipschitz contant of any data-driven algorithm
achieving a desired level of performance has a fundamental lower bound
that depends only on the loss function~$\ell$ and the data-generating
distribution~$\sigma$, which constitutes a fundamental lower bound to
benchmark any training algorithm and learning problem. We also provide
a training scheme for further improving the robustness of a minimizer
with a margin on the loss, by using a graph-based iterative procedure
that involves solving a series of $p$-Poisson equations, decsribed in
Section~\ref{sec: bounds}.

\textbf{Related work.}  Motivated by real-world incidents and
empirical studies~\cite{AK-IG-SB:16b}, the issue of robustness of
data-driven models to adversarial perturbations has received extensive
attention in the last years~\cite{EW-ZK:18, AR-JS-PL:18,
  AI-SS-DT-LE-BT-AM:19b, AF-SMD-PF:16}.  When perturbations are chosen
carefully, early studies~\cite{CS-WZ-IS-JB-DE-IG-RF:14} have shown
that small input variations can cause large prediction errors in
otherwise highly accurate neural networks.
Several frameworks exist to design robust data-driven models,
including regularization~\cite{HG-EF-BP-MC:18}, adversarial
training~\cite{FT-AK-NP-IG-DB-PM:17}, distributionally robust
optimization~\cite{DK-PME-VAN-SSA:19} and training under Lipschitz
constraints. Of the above, the latter approach is particularly
attractive, as it results in trained models with certified~robustness.

The study of robustness of the class of neural network models has
particularly drawn a lot of attention~\cite{PLB-DJF-MJT:17,
  OB-YI-LL-DV-AN-AC:16, LW-HZ-HC-ZS-CJH-LD-DB-ID:18, SZ-YS-TL-IG:16,
  TWW-HZ-PYC-JY-DS-YG-CJH-LD:18, HZ-TWW-PYC-CJH-LD:18,
  JS-RG-GS-MR:17}. Many works~\cite{PP-AK-JB-FA:20, RB-MS-DZ:17,
  CA-JL-RG:18, QL-SH-CA-JL-RG-JHJ:19} explore, in particular, the
problem of training networks with Lipschitz constraints, and related
issues.  The complementary problem of estimating the Lipschitz
constant of a trained neural network is also a crucial part of
providing robustness certificates for trained models, and avoiding the
danger of deploying unsafe models under a false sense of security.
Recent works~\cite{AV-KS:18, MF-AR-HH-MM-GJP:19, PLC-JCP:19} have
focused on deriving upper bounds on the Lipschitz constant of neural
networks.
While these certificates and training schemes provide a way of
estimating and improving robustness of a certain class of data-driven
models, they fall short in providing insight into the~fundamental
robustness bounds for the underlying learning problem and the means to
exploit them in design.

Furthermore, recent works also point towards fundamental tradeoffs
between accuracy and robustness of data-driven
models~\cite{DT-SS-LE-AT-AM:19, AJ-MS-HH:20, AAALM-VK-FP:19b,
  AAALM-VK-FP:19} in various settings and training frameworks.
The connection of adversarial robustness to model complexity and
generalization, and the existence (or non-existence) of fundamental
tradeoffs between them is another important problem that has received
attention~\cite{SG-HW-HY-CY-ZW-JL:19, DS-MH-BS:19,
  SY-KX-SL-HC-JHL-HZ-AZ-KM-YW-XL:19, PN:19, BN-SM-AB-VT-MS-SLJ-IM:18,
  CL-MW-DPK:17}, and is the subject of ongoing debate. This paper
builds and extends upon these early studies.



\textbf{Notation.} We introduce here some useful notation. 
We use~$| \cdot |$ to denote the Euclidean norm in~$\real^d$, for
any~$d \in \mathbb{N}$ (when $d=1$, this denotes the absolute value)
and more generally the Hilbert-Schmidt (H-S) norm in finite
dimensions. We use~$\| \cdot \|$ for function space norms.
For maps~$f$ between high-dimensional spaces, we often require the
notation~$\|~| f |~\|$, which specifies the function space norm
of~$|f|$ (which is in turn the function that evaluates to the H-S norm
of the map~$f$ at any point in its domain).
For~$\mathbb{X} \subset \real^{\dim(\mathbb{X})}$, we denote by
$(\mathbb{X}, \mu)$ the set~$\mathbb{X}$ with an underlying measure~$\mu$.
We denote by~$\mathcal{F}(\mathbb{X}; \mathbb{Y})$ a class~$\mathcal{F}$
(placeholder for the particular spaces mentioned below)
of maps from~$\mathbb{X}$ to~$\mathbb{Y}$.
We denote by~$L^p(\mathbb{X}, \mu)$ the space of~$p$-integrable (measurable) 
functions on~$\mathbb{X}$, where the integration is carried out with the underlying
measure~$\mu$ (the Lebesgue measure is implied when~$\mu$ is not specified),
and by~$W^{1,p}(\mathbb{X}, \mu)$ the space of $p$-integrable (measurable) 
functions with $p$-integrable (measurable) derivatives.
When generalized to the space of maps, as 
in~$f \in L^p((\mathbb{X}, \mu); \mathbb{Y})$, we mean~$|f| \in L^p(\mathbb{X}, \mu)$.
Also, for~$f \in W^p((\mathbb{X}, \mu); \mathbb{Y})$, we mean~$|f| \in L^p(\mathbb{X}, \mu)$
and~$|\nabla f| \in L^p(\mathbb{X}, \mu)$.

\section{Lipschitz-constrained loss minimization and provably robust training}\label{sec: robust_min}
In this section we study and solve the Lipschitz constrained loss
minimization problem~\eqref{eq:adv_robust_expected_loss_min}.
We start by specifying the setting for
Problem~\eqref{eq:adv_robust_expected_loss_min}.
Let $\mathbb{X} \subset \real^{\dim(\mathbb{X})}$
and~$\mathbb{Y} \subset \real^{\dim(\mathbb{Y})}$ be convex and
compact, $\sigma$ an absolutely continuous probability measure
on~$\mathbb{X} \times \mathbb{Y}$ with (absolutely continuous)
marginal~$\mu$ supported on~$\mathbb{X}$ and
conditional~$\pi$. 
%
Let the loss
function~$\ell: \mathbb{Y} \times \mathbb{Y} \rightarrow
\realnonnegative$ be strictly convex and Lipschitz continuous.
The Lipschitz constraint on the maps
in~\eqref{eq:adv_robust_expected_loss_min} is a global constraint
involving every pair of points in the domain~$\mathbb{X}$. To obtain a
tractable formulation, we equivalently rewrite the Lipschitz
constraint as a bound on the norm of the gradient in the
domain~$\mathbb{X}$.
The space of Lipschitz continuous maps~$\Lip(\mathbb{X};\mathbb{Y})$
is also the Sobolev space~$W^{1, \infty}((\mathbb{X},\mu);\mathbb{Y})$
of essentially bounded (measurable) maps with essentially bounded
(measurable) gradients, that is,
$\Lip(\mathbb{X};\mathbb{Y}) = W^{1,
  \infty}((\mathbb{X},\mu);\mathbb{Y})$.\footnote{We let $\mu$ be the
  underlying measure on~$\mathbb{X}$, since the input data is
  generated from $\mu$ on the support $\mathbb{X}$.}
%
The Lipschitz constant of a map~$f \in \Lip(\mathbb{X};\mathbb{Y})$ is
$\lip(f) = \| |\nabla f| \|_{L^{\infty}((\mathbb{X},\mu);\mathbb{Y})}$
(the~$W^{1,\infty}$-seminorm of~$f$). We refer the reader to our
supplementary material or~\cite{LCE:98} for a discussion of these
notions.

Using the above definitions, the Lipschitz constrained loss
minimization problem~\eqref{eq:adv_robust_expected_loss_min} becomes
\begin{align}\label{eq:lip_constrained_loss_min}
  \inf_{f \in W^{1,\infty}((\mathbb{X},\mu);\mathbb{Y})} \left \lbrace
  L_{\sigma}(f), \qquad
  \text{s.t.}~~\left\| |  \nabla f | \right\|_{L^{\infty}(\mathbb{X},
  \mu)} \leq \alpha \right \rbrace.
\end{align}
%
%
To see the role of the Lipschitz constant in the 
sensitivity of the loss to adversarial perturbations,
first notice that adversarial perturbations
can be written as the perturbations on the joint distribution~$\sigma$
generated by a map~$T$ that perturbs the
inputs~$x \in \mathbb{X}$ while preserving the
outputs~$y \in \mathbb{Y}$~\cite{CS-WZ-IS-JB-DE-IG-RF:14}. 
In compact form, the class of adversarial
perturbations can be written as:
\begin{align*}
  \mathcal{T} = \left \lbrace T 
  \; \left| \; T(x,y) = (T_1(x,y) \; , \; y),~ \text{s.t.}~~T_1(x,y)
  \in B_{\delta}(x) \cap \mathbb{X} \right. \right \rbrace,
\end{align*}
where~$B_{\delta}(x)$ is the open ball in~$\real^{\dim(\mathbb{X})}$
of radius~$\delta > 0$ and centered at~$x$. Defining the sensitivity as
the worst-case increase of the loss~$L_{\sigma}$ following an
adversarial perturbation~$T \in \mathcal{T}$ for any~$\sigma$, we
get\footnote{See Supplementary Material for a proof.} that it is modulated by~$L^{\infty}$-norm of the 
gradient~$\nabla_1 \ell \cdot \nabla f$ (precisely,
$\| | \nabla_1 \ell \cdot \nabla f | \|_{L^{\infty}(\mathbb{X} \times
  \mathbb{Y}, \sigma)}$)\footnote{We use $\nabla_1 \ell$ to denote
  the gradient of $\ell$ with respect to its first
  argument.} and whose upper bound is 
 determined by the Lipschitz constant:
\begin{align}\label{eq: inequality derivative Lipschitz}
  \underbrace{\| | \nabla_1 \ell \cdot \nabla f | \|_{L^{\infty}(\mathbb{X} \times
  \mathbb{Y}, \sigma)}}_{\text{sensitivity of $L$ to adv. perturbation}} \leq \underbrace{\| | \nabla_1 \ell |
  \|_{L^{\infty}(\mathbb{X} \times \mathbb{Y},
  \sigma)}}_{\text{Lipschitz constant of $\ell$}}  \cdot
  \underbrace{\| | \nabla f | \|_{L^{\infty}(\mathbb{X},
  \mu)}}_{\text{Lipschitz constant of $f$}}.
\end{align}
%
%

Problem~\ref{eq:lip_constrained_loss_min} is convex (owing to the
strict convexity of the loss~$L_{\sigma}$\footnote{See supplementary
  material for a proof.}
and the convexity of the
constraint). Thus, we can expect to obtain a (unique) minimizer from the
saddle point of the corresponding Lagrangian.
With
$G_f(x) = \frac{1}{2} \left( | \nabla f (x) |^2 - \alpha^2 \right)$,
we can reformulate the Lipschitz constraint as $G_f \leq 0$
$\mu$--a.e. in~$\mathbb{X}$\footnote{The constraint violation set is of zero measure, that is, $\mu \left( \lbrace x \in \mathbb{X} \; | \; G_f(x) > 0 \rbrace \right) = 0$.}.
Since~$f \in W^{1,\infty}((\mathbb{X},\mu);\mathbb{Y})$, the
constraint function~$G_f$ belongs to the
space~$ L^{\infty}(\mathbb{X}, \mu)$.  Correspondingly, the Lagrange
multiplier for the constraint $G_f \leq 0$
($\mu$--a.e. in~$\mathbb{X}$) is non-negative\footnote{Any
  $\lambda \in L^{\infty}(\mathbb{X}, \mu)^*$ is also a bounded,
  finitely additive (absolutely continuous) measure on~$\mathbb{X}$.}
%
and belongs to the dual space of $L^{\infty}(\mathbb{X}, \mu)$, which
we denote as~$\lambda \in L^{\infty}(\mathbb{X}, \mu)^*_{\geq 0}$.
The
Lagrangian~$\mathcal{L}_{\sigma} :
W^{1,\infty}((\mathbb{X},\mu);\mathbb{Y}) \times
L^{\infty}(\mathbb{X}, \mu)^*_{\geq 0}$ for
Problem~\eqref{eq:lip_constrained_loss_min} is then given by:
\begin{align}\label{eq: Lagrangian problem 2}
  \mathcal{L}_{\sigma}(f, \lambda) = L_{\sigma}(f) + \lambda \left( G_f \right).
\end{align}
\begin{theorem}\textbf{\textit{(Lipschitz constrained loss
      minimization)}}\label{thm:saddle_point_Lip_constraint}
Problem~\eqref{eq:lip_constrained_loss_min} has a unique global 
minimizer~$f^* \in W^{1,\infty}((\mathbb{X},\mu);\mathbb{Y})$.
%
The Lagrangian~$\mathcal{L}_{\sigma}$ has a unique saddle point 
$(f^*, \lambda^*) \in W^{1,\infty}((\mathbb{X},\mu);\mathbb{Y}) \times L^1(\mathbb{X}, \mu)_{\geq 0}$.
Moreover,
$(f^*, \lambda^*)$ satisfies the
first-order optimality conditions:
\begin{enumerate}
\item \textit{Stationarity:} 
The saddle point~$(f^*, \lambda^*)$ is a weak solution of the Poisson equation,
\begin{align}
\begin{aligned}
	- \frac{1}{\mu} \nabla \cdot \left(\mu \lambda^* \nabla f^* \right) + g_{f^*} = 0 ~~~ \text{in}~ \mathbb{X}, \qquad
				 \mu \lambda^* \nabla f^* \cdot \mathbf{n} = 0 ~~~ \text{on}~\partial \mathbb{X},
\end{aligned}
		\label{eq:saddle_point_lagrangian_lip_constrained_loss_min}
\end{align}
where $g_{f^*}(x) = \mathbb{E}_{y \sim \pi(y \; | \; x)} \left[ \nabla_1 \ell(f^*(x), y) \right]$
and~$\mathbf{n}$ is the outward normal to the boundary~$\partial \mathbb{X}$.
\item \textit{Feasibility:}
$|\nabla f^*| \leq \alpha$ and $\lambda^* \geq 0$, $\mu-\text{a.e. in}~\mathbb{X}$.
\item \textit{Complementary slackness:}
$\lambda^* \left( |\nabla f^*| - \alpha \right) =0$, $\mu-\text{a.e. in}~\mathbb{X}$.
\end{enumerate}
\end{theorem}
%
%
Some comments on Theorem \ref{thm:saddle_point_Lip_constraint} are in
order. In the absence of the constraint
in~\eqref{eq:lip_constrained_loss_min} (that is, $\alpha = \infty$),
the stationarity condition is characterized
by~$\mathbb{E}_{y \sim \pi(y \; | \; x)} \left[ \nabla_1
  \ell(f^*_{\text{unc}}(x), y) \right] = 0$,
where~$f^*_{\text{unc}}(x), y)$ is the unconstrained minimizer of the
loss functional.
The saddle point of~$\mathcal{L}_{\sigma}$ is characterized by the
Poisson
equation~\eqref{eq:saddle_point_lagrangian_lip_constrained_loss_min},
which encodes the stationarity condition for the Lagrangian.  The
Neumann boundary condition
in~\eqref{eq:saddle_point_lagrangian_lip_constrained_loss_min} results
from the fact that we do not enforce a boundary constraint on the map
in the loss minimization problem~\eqref{eq:lip_constrained_loss_min}.
The $\lambda^*$-weighted Laplace operator,
$\frac{1}{\mu} \nabla \cdot \left(\mu \lambda^* \nabla \right)$, is
responsible for locally enforcing the Lipschitz constraint and
regularizing (smoothing) the minimizer.  Moreover, the Lagrange
multiplier satisfies~$\lambda^* \in L^1(\mathbb{X}, \mu)_{\geq 0}$,
and is therefore integrable (this is stronger regularity than in the
definition~$\lambda \in L^{\infty}(\mathbb{X}, \mu)^*_{\geq 0}$).
%
%
It follows from the feasibility condition in
Theorem~\ref{thm:saddle_point_Lip_constraint} that the minimizer
(provably) satisfies the Lipschitz bound (in contrast to Lipschitz
regularization-based approaches to adversarial learning).
From the complementary slackness condition in
Theorem~\ref{thm:saddle_point_Lip_constraint}, smoothing is enforced
only when the constraint is active: when the constraint is inactive in
a region~$D \subset \mathbb{X}$ of non-zero measure (that is,
$|\nabla f^*(x)| < \alpha$ for~$x \in D$ and~$\mu(D) > 0$), the
Lagrange multiplier satisfies~$\lambda^* = 0$ ($\mu$-a.e.~in~$D$) and
smoothing is not enforced.


The fact that the saddle point of the
Lagrangian~$\mathcal{L}_{\sigma}$ in~\eqref{eq: Lagrangian problem 2}
satisfies the Lipschitz bound forms the basis for the design of a
provably robust training scheme, which we obtain through a
discretization of Problem~\eqref{eq:lip_constrained_loss_min} over a
graph. To this end, we select~$n$ points
$\left \lbrace X_i \right \rbrace_{i=1}^n$, $X_i \in \mathbb{X}$, via
i.i.d.~sampling of the distribution~$\mu$ (in practice, we sample
uniformly i.i.d.~from the input dataset, that defines the empirical
marginal measure~$\widehat{\mu}$). With the discretization points
$\left \lbrace X_i \right \rbrace_{i=1}^n$ as the (embedding of)
vertices, we construct an undirected, weighted, connected
graph~$\mathcal{G} = (\mathcal{V}, \mathcal{E}, W )$, with vertex
set~$\mathcal{V} = \lbrace 1, \ldots, n \rbrace$, edge
set~$\mathcal{E} = \mathcal{V} \times \mathcal{V}$, and weighted
adjacency matrix~$W = [w_{ij}]_{i,j = 1}^n$.

We assume the availability of a labeled dataset
  $D = \lbrace (x_i, y_i) \rbrace_{i=1}^N$ consisting of ~$N > n$
  i.i.d.~samples of~$\sigma$,
and define a partition~$\mathcal{W} = \lbrace \mathcal{W}_i \rbrace_{i=1}^n$ of the
dataset $D$ as follows:
\begin{align}\label{eq:dataset_partition}
  \mathcal{W}_i = \left \lbrace  (x,y) \in D \; | \;  \left| x - X_i
  \right| \leq \left| x - X_j \right| ~\forall~ j \in \mathcal{V}
  \setminus \lbrace i \rbrace  \right \rbrace .
\end{align}
We then assign weights $\theta_{ij} = N^{-1}$ to the
samples~$\xi_j = (x_j, y_j) \in \mathcal{W}_i$ (a different weighing
scheme may affect generalization and performance of our model; we
leave this for future research).
Finally, we write the discrete (empirical) Lipschitz constrained loss
minimization problem over the graph~$\mathcal{G}$ as follows (this
minimization problem can be viewed as the discretized version
of~\eqref{eq:lip_constrained_loss_min} over $\mathcal{G}$):
\begin{align}\label{eq:empirical_loss_min_graph_discretization}
  \min_{ \substack{\mathbf{v} = (v_1, \ldots, v_n) \\ v_i \in \real^{\dim(\mathbb{Y})} } }
  ~\left \lbrace \sum_{i \in \mathcal{V}} \left( \sum_{j \in
  \mathcal{W}_i}  \theta_{ij} \ell ( v_i , y_j ) \right),
  ~~~ \text{s.t.}~~ \left| v_r - v_s \right| \leq \alpha \left| X_r -
  X_s \right|, ~~\forall~(r,s) \in \mathcal{E} \right \rbrace.
\end{align}
We note that the above constrained minimization
problem~\eqref{eq:empirical_loss_min_graph_discretization} is convex
(strictly convex objective function with convex constraints) and the
corresponding Lagrangian is given by:
\begin{align}
	\mathcal{L}_{\mathcal{G}}(\mathbf{v}, \Lambda) = \sum_{i \in \mathcal{V}} \left[ \sum_{s \in \mathcal{W}_i} \theta_{is} \ell ( v_i, y_s )  
	+  \frac{1}{2} \sum_{j \in \mathcal{V}} \lambda_{ij} w_{ij} \left( \left| v_i - v_j \right|^2 - \alpha \left| X_i - X_j \right|^2 \right) \right],
\end{align}
where~$\Lambda = [\lambda_{ij}]_{i,j=1}^n$ is the matrix of Lagrange
multiplier for the pairwise Lipschitz constraints. Define a
primal-dual dynamics for the
Lagrangian~$\mathcal{L}_{\mathcal{G}}(\mathbf{v}, \Lambda)$ with
time-step sequence~$\lbrace h(k) \rbrace_{k \in \mathbb{N}}$:
\begin{align}\label{eq:p-d_Lipschitz_const_loss_min}
\begin{aligned}
	\mathbf{v}(k+1) &= \mathbf{v}(k) - h(k) ~\nabla_{\mathbf{v}} \mathcal{L}_{\mathcal{G}} \left( \mathbf{v}(k)  ,  \Lambda(k) \right), \\
	 \Lambda(k+1) &= \max \lbrace 0\; , \; \Lambda(k) + h(k) ~\nabla_{\Lambda} \mathcal{L}_{\mathcal{G}} \left(\mathbf{v}(k)  ,  \Lambda(k) \right) \rbrace.
\end{aligned}
\end{align}
The primal dynamics is a discretized heat flow over the
graph~$\mathcal{G}$ with a weighted Laplacian,
where~$\nabla_{\mathbf{v}} \mathcal{L}_{\mathcal{G}} \left(
  \mathbf{v}(k) , \Lambda(k) \right) = \left( \Delta(\Lambda, W)
  \otimes I_{\dim(\mathbb{Y})} \right) \mathbf{v} + \theta \cdot
\nabla_1 \ell (\mathbf{v}, \mathbf{y})$, and~$\Delta(\Lambda, W)$ is
the $\Lambda \circ W$-weighted Laplacian of the graph~$\mathcal{G}$
(where~$\circ$ denotes the Hadamard or entry-wise product of
matrices).  The convergence of the
solution~$\lbrace (\mathbf{v}(k), \Lambda(k)) \rbrace_{k \in
  \mathbb{N}}$ of the primal-dual
dynamics~\eqref{eq:p-d_Lipschitz_const_loss_min} to the saddle point
of the Lagrangian~$\mathcal{L}_{\mathcal{G}}$
follows~\cite{KA-HA-LH-HU:58} from the convexity of
Problem~\eqref{eq:empirical_loss_min_graph_discretization}.

As the size of the dataset $N$ and the size of graph $n$ increase, the
solution to Problem~\eqref{eq:empirical_loss_min_graph_discretization}
approaches the solution to Problem~\eqref{eq:lip_constrained_loss_min}, 
under certain mild conditions.
In particular, by the Glivenko-Cantelli
Theorem~\cite{PB:08}, the empirical
measure~$\widehat{\sigma}_N = \frac{1}{N} \sum_{i=1}^N
\delta_{(x_i,y_i)}$
converges uniformly and almost surely to the distribution~$\sigma$ in
the limit for~$N \rightarrow \infty$, and so
does~$\widehat{\mu}_n = \frac{1}{n} \sum_{i=1}^n \delta_{X_i}
\rightarrow \mu$ as~$n \rightarrow \infty$, where~$\delta$ here
denotes the Dirac measure. Further, the convergence
as~$n \rightarrow \infty$ (higher model complexity) and
$N \rightarrow \infty$ (larger dataset) of the minimizer of the
(empirical) discrete minimization
problem~\eqref{eq:empirical_loss_min_graph_discretization} to the
infinite-dimensional problem~\eqref{eq:lip_constrained_loss_min} is
modulated by the weights~$\theta$ (which govern the convergence of the
empirical loss) and~$w$ (which governs the convergence of the graph Laplacian to the Laplace operator
on the domain~\cite{MB-PN:08}). 

\begin{figure}[tb]
  \centering
    \includegraphics[width=1\columnwidth]{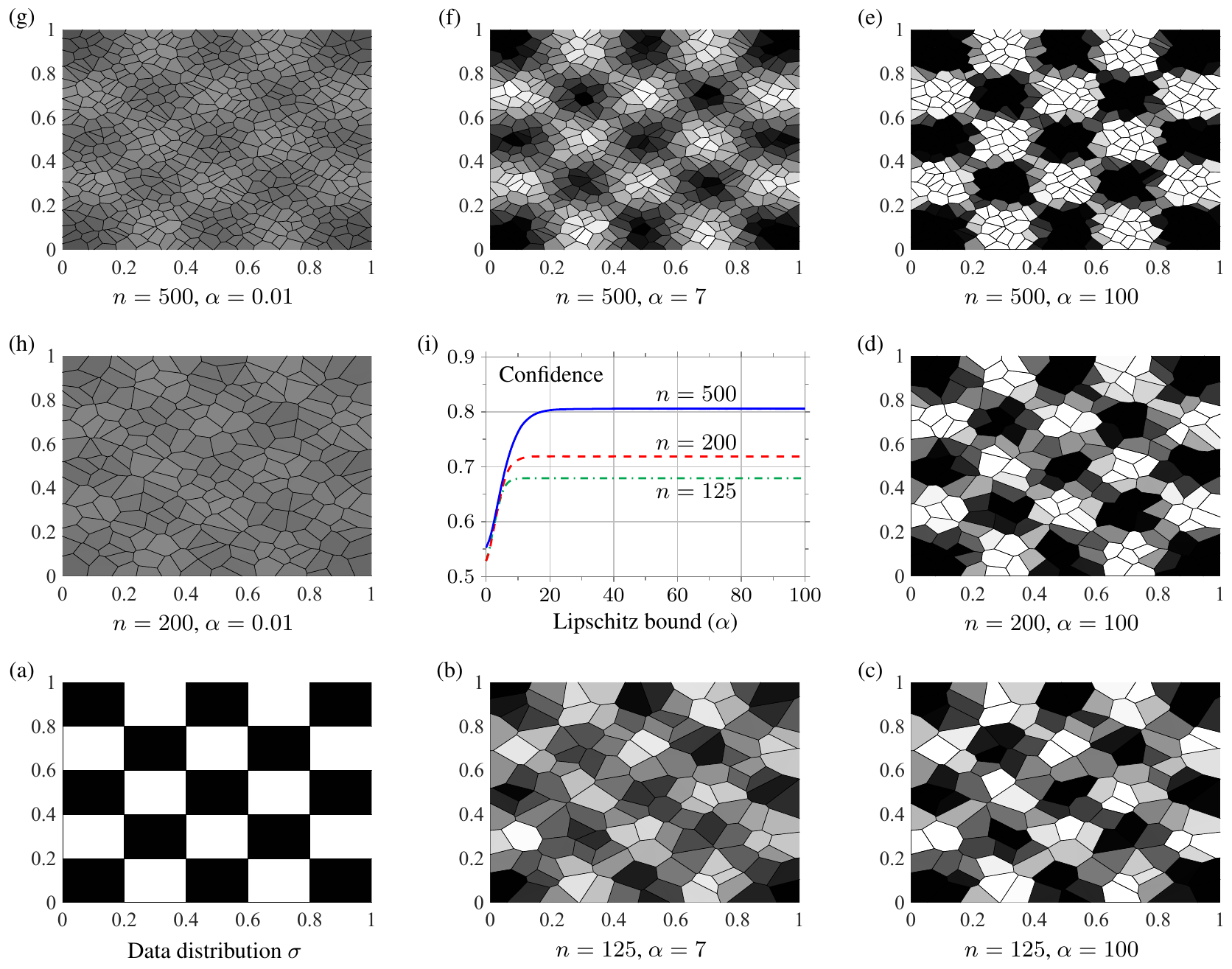} 
    \caption[]{For the classification problem discussed in Section
      \ref{sec: robust_min}, this figure shows a tradeoff between the
      confidence of classification, the Lipschitz constant, and the
      complexity of the classifier designed using our algorithm
      \eqref{eq:p-d_Lipschitz_const_loss_min}. Increasing the
      Lipschitz constant of the classifier and its complexity also
      increases the confidence of classification, at the expenses of a
      higher sensitivity to perturbations.}
  \label{fig:feedbackInterconnection}
\end{figure} 

We conclude this section with an illustrative example. Consider a
dataset of $10000$ i.i.d. samples $(x_i,y_i)$, with $x_i \in [0,1]^2$
and $y_i \in \{[1 \;0]^\transpose, [0 \; 1]^\transpose \}$, taken
uniformly from the distribution $\sigma$ in
Fig.~\ref{fig:feedbackInterconnection}(a), where
$y_i = [1 \;0]^\transpose$ if $x_i$ belongs to a white cell and
$y_i = [0 \; 1]^\transpose$ if $x_i$ belongs to a black cell. We
randomly select $n$ nodes in $[0,1]^2$, with $n = \{125, 200, 500\}$,
construct a graph $\mc G= (\mc V, \mc E)$ by connecting each node to
its $10$ nearest neighbors, and compute the solution $\mathbf{v}^*$ to
\eqref{eq:p-d_Lipschitz_const_loss_min} for different values of the
Lipschitz constant $\alpha$. Then, we generate a testing set of $2000$
i.i.d. samples from $\sigma$, associate them with the closest node,
and evaluate the classification confidence of $\mathbf{v}^*$. In
particular, if the testing sample $\bar x_i$ is closest to the $i$-th
node and $v_i^* = [p_1 \; p_2]^\transpose$, then $\bar x_i$ is
classified as $[1 \; 0]^\transpose$ with confidence $p_1$ if
$p_1 > p_2$, and as $[0 \; 1]^\transpose$ with confidence $1-p_1$ if
$p_1 < p_2$. Fig. \ref{fig:feedbackInterconnection}(b)-(h) shows the
Voronoi cells associated with the nodes $\mc V$, where each cell is
colored on a gray scale using the first entries of $v_i^*$ (darker
colors indicate higher confidence in classifying the samples in a cell
as $[0 \; 1]^\transpose$, while lighter colors indicate higher
confidence in classifying the samples in a cell as
$[1 \; 0]^\transpose$). It can be seen that the classification
confidence increases with the number of nodes and the Lipschitz bound,
at the expenses of a higher model complexity and sensitivity to
adversarial perturbations. This trend is also visible in
Fig.\ref{fig:feedbackInterconnection}(i), where the classification
confidence increases with the Lipschitz bound until it saturates for
the classifier with highest confidence given the training set and
discretization points.

\section{Robustification with loss margin and fundamental bound}\label{sec: bounds}
In this section we study the problem of increasing the robustness of a
minimizer with a margin on the loss. Let $f^*$ be the minimizer of
\eqref{eq:adv_robust_expected_loss_min} with Lipschitz bound $\alpha$,
and let $J^*_{\sigma}(\alpha)$ be the optimal loss. We formulate and
solve the following loss constrained Lipschitz constant minimization
problem:
\begin{align}\label{eq:lipschitz_min}
  \inf_{f \in  W^{1, \infty}((\mathbb{X},\mu);\mathbb{Y})} \left
  \lbrace \| |\nabla f| \|_{L^{\infty}((\mathbb{X},\mu);\mathbb{Y})}, 
  \qquad \text{s.t.}~~ L_{\sigma}(f) \leq J^*_{\sigma}(\alpha) +
  \epsilon \right \rbrace.
\end{align}
Because the Lipschitz constant satisfies
$\| |\nabla f| \|_{L^{\infty}((\mathbb{X},\mu);\mathbb{Y})} = \ess
\sup |\nabla f|$, Problem~\eqref{eq:lipschitz_min} has a $\min - \max$
(more precisely, an~$\inf - \ess \sup$) structure which is not
amenable to tractable numerical schemes. We circumvent this hurdle by
approaching problem~\eqref{eq:lipschitz_min} via a sequence of
loss-constrained (convex) minimization problems involving
the~$W^{1,p}$-seminorm, for~$p \in \mathbb{N}$,~$p>1$, given by:
\begin{align}\label{eq:W^1,p_min}
  \inf_{f \in W^{1,p}((\mathbb{X},\mu);\mathbb{Y})} \left \lbrace
  \left \| | \nabla f| \right \|_{L^p(\mathbb{X},\mu)} , 
  \qquad \text{s.t.}~~ L_{\sigma}(f) \leq J^*_{\sigma}(\alpha) +
  \epsilon \right \rbrace.
\end{align}
$W^{1,p}$-seminorm minimization problems are typically formulated to
obtain minimum Lipschitz extensions in semi-supervised
learning~\cite{AEA-XC-AR-MW-MIJ:16, RK-AR-SS-DAS:15, RKA-TZ:07,
  JC:19}.  A related problem is the one of $W^{1,p}$-seminorm
regularized learning~\cite{ALB-AF:12, EM-TK-ALB:13}. Instead, we
propose this approach, for the first time, to improve the robustness
of minimizers to adversarial perturbations with a guaranteed margin on
the loss.

Convexity of Problem~\eqref{eq:W^1,p_min} follows from the convexity
of the~$W^{1,p}$-seminorm in~$W^{1,p}((\mathbb{X},\mu);\mathbb{Y})$
and the strict convexity of~$L_{\sigma}$ (which yields a convex
constraint). The minimizers are obtained from the saddle points of the
Lagrangian~$\mathcal{H}^p_{\sigma} :
W^{1,p}((\mathbb{X},\mu);\mathbb{Y}) \times \realnonnegative
\rightarrow \real$ for Problem~\eqref{eq:W^1,p_min}, given by:
\begin{align}
	\mathcal{H}^p_{\sigma}(f, \kappa) = \frac{1}{p} \left \|  |\nabla f|  \right \|^p_{L^p(\mathbb{X}, \mu)} + \kappa \left( L_{\sigma}(f) -(J^*_{\sigma}(\alpha) + \epsilon) \right),
	\label{eq:Lagrangian_H_Lip_min}
\end{align}
where we (equivalently) consider the~$p$-th
exponent~$\left \| |\nabla f| \right \|^p_{L^p(\mathbb{X}, \mu)}$ of
the~$W^{1,p}$-seminorm in defining the Lagrangian. 
The saddle points of~$\mathcal{H}^p_{\sigma}$ are now specified by a
Poisson equation involving the~$p$-Laplace operator,\footnote{The
  $p$-Laplace operator is defined as
  $\Delta^{\mu}_p u = \frac{1}{\mu} \nabla \cdot \left( \mu |\nabla
    u|^{p-2} \nabla u \right)$.} as established in the following
theorem:
\begin{theorem}\textbf{\textit{(Loss constrained $W^{1,p}$-seminorm
      minimization)}}\label{thm:saddle_point_W^1,p_min}
  For every~$p \in \mathbb{N}_{>1}$, there exists a global
  minimizer~$f^{\epsilon, p} \in W^{1,p}((\mathbb{X},\mu);\mathbb{Y})$
  for Problem~\eqref{eq:W^1,p_min}. Also, there exists a
  saddle
  point~$(f^{\epsilon, p}, \kappa^{\epsilon, p}) \in
  W^{1,p}((\mathbb{X},\mu);\mathbb{Y}) \times \realnonnegative$ of the
  Lagrangian~$\mathcal{H}_{\sigma}^p$.
  Moreover,~$(u, \kappa) \in W^{1,p}((\mathbb{X},\mu);\mathbb{Y})
  \times \realnonnegative$ is a saddle point
  of~$\mathcal{H}^p_{\sigma}$ if and only if it satisfies the
  following first-order optimality conditions:
  \begin{enumerate}
  \item \textit{Stationarity:} $(u, \kappa)$ is a (weak) solution of
    the $p$-Poisson equation:
    \begin{align}\label{eq:W^1,p_saddle_Poisson}
      - \Delta^{\mu}_p u + \kappa  g_u = 0~~\text{in}~ \mathbb{X}, 
      \qquad  \mu \nabla u \cdot \mathbf{n} = 0 ~~\text{on}~\partial \mathbb{X},
    \end{align}
    where~$g_u(x) = \mathbb{E}_{y \sim \pi(y \; | \; x)} \left[
      \nabla_1 \ell(u(x), y) \right]$ and~$\Delta^{\mu}_p$ is the
    $p$-Laplace operator on~$(\mathbb{X}, \mu)$.
    
  \item \textit{Feasibility:} $L_{\sigma}(u) \leq
    J^*_{\sigma}(\alpha) + \epsilon$ and~$\kappa \geq 0$.

  \item \textit{Complementary slackness:}
    $\kappa \left( L_{\sigma}(f) -(J^*_{\sigma}(\alpha) + \epsilon)
    \right) = 0$.
\end{enumerate}
\end{theorem}
With the characterization of the minimizers of~\eqref{eq:W^1,p_min}
for every~$p \in \mathbb{N},~p>1$ from
Theorem~\ref{thm:saddle_point_W^1,p_min}, we now investigate whether
the minimum value of~\eqref{eq:W^1,p_min} and its minimizers converge
(as~$p \rightarrow \infty$) to those of~\eqref{eq:lipschitz_min}. The
following theorem establishes that this is indeed the case, and that
the minimum Lipschitz constant in~\eqref{eq:lipschitz_min} can be
obtained as the limit of the sequence of minimum values
of~\eqref{eq:W^1,p_min}.
\begin{theorem}\textbf{\textit{(Limit as~$p \rightarrow \infty$ and
      fundamental Lipschitz lower bound)}}\label{thm:conv_W^1,p_Lip}
  For any~$\epsilon > 0$, it holds
  \begin{align*}
    \lim_{p \rightarrow \infty} ~ \min_{\substack{f \in
    W^{1,p}((\mathbb{X},\mu);\mathbb{Y})  \\
    L_{\sigma}(f) \leq J^*_{\sigma}(\alpha) + \epsilon}} ~ \left \| |
    \nabla f| \right \|_{L^p(\mathbb{X},\mu)}
    = \min_{\substack{f \in W^{1, \infty}((\mathbb{X},\mu);\mathbb{Y})
    \\
    L_{\sigma}(f) \leq J^*_{\sigma}(\alpha) + \epsilon }} ~\| |\nabla
    f| \|_{L^{\infty}((\mathbb{X},\mu);\mathbb{Y})}.
\end{align*}
Moreover, as~$p \rightarrow \infty$,
the sequence~$\left \lbrace f^{\epsilon, p} \right \rbrace_{p \in
\mathbb{N}_{>1}}$ of minimizers of Problem~\eqref{eq:W^1,p_min}
converges uniformly to a (global) minimizer~$f^{\epsilon, \infty}$
of~\eqref{eq:lipschitz_min}.
\end{theorem}

%

%

The facts that the saddle points of~$\mathcal{H}_{\sigma}^p$
in~\eqref{eq:Lagrangian_H_Lip_min} satisfy the bound on the loss (for
every~$p \in \mathbb{N}_{>1}$) for a given margin~$\epsilon > 0$, and
that the minimum value and minimizers of~\eqref{eq:W^1,p_min} converge
in the limit~$p \rightarrow \infty$ to those
of~\eqref{eq:lipschitz_min}, form the basis for the design of a
robustification scheme.
With the same graph structure and dataset partitioning as in
Section~\ref{sec: robust_min}, we write the discrete (empirical)
loss-constrained $W^{1,p}$-seminorm minimization problem over the
graph~$\mathcal{G}$ as follows (this minimization problem can be
viewed as the discretized version of~\eqref{eq:W^1,p_min} over the
structure imposed by~$\mathcal{G}$):
\begin{align}\label{eq:W^1,p_min_graph_discretization}
  \min_{\substack{\mathbf{v} = (v_1, \ldots, v_n) \\
  v_i \in \real^{\dim(\mathbb{Y})}}} 
  \left \lbrace \frac{1}{p} \sum_{i \in \mathcal{V}} \sum_{j \in
  \mathcal{N}_i} w_{ij}  \left| v_i - v_j \right|^p  ,
  \qquad \text{s.t.}~~ \sum_{i \in \mathcal{V}} \sum_{s \in
  \mathcal{W}_i} \theta_{is} \ell ( v_i , y_s ) \leq
  J^*_{\sigma}(\alpha) + \epsilon  \right \rbrace.   
\end{align}
We note that the above constrained minimization
problem~\eqref{eq:W^1,p_min_graph_discretization} is convex (convex
objective function with convex constraints), and that the
corresponding Lagrangian is given by:
\begin{align}\label{eq:discrete_lagrangian_W^1,p_graph}
  \mathcal{H}_{\mathcal{G}}^p (\mathbf{v}, \kappa) 
  = \sum_{i \in \mathcal{V}}  \left[ \frac{1}{p} \sum_{j \in
  \mathcal{N}_i} w_{ij} \left| v_i - v_j \right|^p 
  + \kappa \sum_{s \in \mathcal{W}_i} \left( \theta_{is} \ell ( v_i,
  y_s ) - \frac{1}{n}(J^*_{\sigma}(\alpha) + \epsilon) \right) \right],
\end{align} 
The saddle points of \eqref{eq:discrete_lagrangian_W^1,p_graph} can be
obtained via a primal-dual algorithm similar
to~\eqref{eq:p-d_Lipschitz_const_loss_min} in Section~\ref{sec:
  robust_min}. We solve the (discrete) loss-constrained Lipschitz
minimization problem using an iterative procedure that employs the
primal-dual algorithm to converge to a saddle point
of~$\mathcal{H}_{\mathcal{G}}^p$
in~\eqref{eq:discrete_lagrangian_W^1,p_graph} at every iteration
step~$p \in \mathbb{N}_{>1}$.  We then use the saddle point
of~$\mathcal{H}_{\mathcal{G}}^p$ as the initialization for the
iteration step~$p+1$.

Theorem~\ref{thm:saddle_point_W^1,p_min} offers key insights on the
fundamental tradeoff between robustness and nominal performance.  From
complementary slackness in Theorem~\ref{thm:saddle_point_W^1,p_min},
it follows that, for the saddle
points~$(f^{\epsilon, p}, \kappa^{\epsilon, p})$, either the Lagrange
multiplier satisfies~$\kappa^{\epsilon, p} = 0$ or the constraint is
active ($f^{\epsilon, p}$ occurs at the boundary of the constraint and
the loss is
$L_{\sigma}(f^{\epsilon, p}) = J^*(\alpha) + \varepsilon$).  If the
Lagrange multiplier is zero, then the Poisson equation characterizing
the Stationarity condition~\eqref{eq:W^1,p_saddle_Poisson} reduces to
the $p$-Laplace equation with a Neumann boundary condition, whose
solution is a constant map (in the weak sense).  However, in
practically useful cases (for small values of~$\alpha$
and~$\varepsilon$) with a low optimal loss~$J^*(\alpha)$, there will
typically not exist a constant map satisfying the loss
margin~$\varepsilon$ (unless the unconstrained
minimizer~$f^*_{\text{unc}}$ is itself flat). This implies that the
Lagrange multiplier~$\kappa$ is typically nonzero, that the
minimizer~$f^{\epsilon, p}$ occurs at the constraint boundary, and
that the loss
satisfies~$L_{\sigma}((f^{\epsilon, p}) = J^*(\alpha) + \varepsilon$.
Therefore, for every $p \in \mathbb{N}_{>1}$, the minimization
problem~\eqref{eq:W^1,p_min} is typically dominated by the constraint,
and the minimum value of the~$W^{1,p}$-norm decreases monotonically
with the loss margin. Thus, a fundamental tradeoff exists between
performance and~robustness.

\begin{figure}[tb]
  \centering
    \includegraphics[width=1\columnwidth]{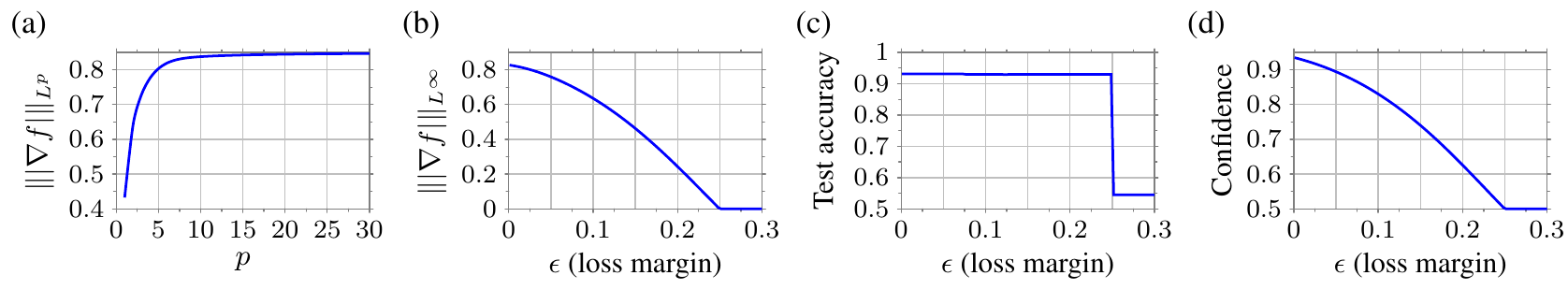} 
    \caption[]{For the classification problem discussed in Section
      \ref{sec: robust_min}, (a) shows the convergence of the minimum
      values of~\eqref{eq:W^1,p_min_graph_discretization}
      as~$p \rightarrow \infty$ to the
        minimum Lipschitz constant (b) shows the tradeoff between
      performance and robustness (monotonic decrease of the minimum
      Lipschitz constant as a function of the loss margin), 
      and (c)-(d) show the dependence of accuracy and confidence in testing on the loss margin
      obtained by solving \eqref{eq:W^1,p_min_graph_discretization}
      for different values of $\varepsilon$.}
  \label{fig:lipschitz_vs_loss_margin} 
\end{figure} 

We conclude this section with an example. Consider the classification
problem described in Section \ref{sec:
  robust_min}. Fig. \ref{fig:lipschitz_vs_loss_margin} shows the
properties of the minimizers to
\eqref{eq:W^1,p_min_graph_discretization} for varying values of $p$
and~$\varepsilon$. It can be seen that, (i) as $p$ increases, the
minimum value of~\eqref{eq:W^1,p_min_graph_discretization} converges
to its supremum value, which, by Theorem~\ref{thm:conv_W^1,p_Lip}, is
smallest Lipschitz constant for a guaranteed loss margin $\varepsilon$
(Fig. \ref{fig:lipschitz_vs_loss_margin}(a)), and (ii) the minimum
Lipschitz constant associated with the loss-constrained minimization
problem is a monotonically non-increasing function of the loss margin
$\varepsilon$, and strictly decreasing for small values of $\alpha$
and $\varepsilon$ (Fig. \ref{fig:lipschitz_vs_loss_margin}(b)). This
curve describes a fundamental tradeoff between adversarial robustness
and performance, and is entirely determined by the properties of the
classification problem and not by the structure of the~classifier. 
Fig. \ref{fig:lipschitz_vs_loss_margin}(c) and (d) show the dependence of accuracy and confidence in testing
on the loss margin in training, and as expected, 
they are decreasing functions of the loss margin. 
We observe in (c) that the accuracy is constant at $0.93$ for $\epsilon<0.25$ then drops to $0.54$ at $\epsilon=0.25$. 
On the other hand, we observe that the confidence in (d) decreases smoothly with the loss margin for $\epsilon<0.25$ till it reaches $0.5$ at $\epsilon = 0.25$. 
This implies that although the testing accuracy of the classifier remains at $0.93$ for $\epsilon<0.25$, 
the classification is made with progressively lower confidence. 
For $\epsilon \geq 0.25$, the accuracy of the classifier is $0.54$
while the classification is made with a confidence of $0.5$ for each
of the two classes.

\section{Numerical experiments on MNIST dataset}\label{sec: mnist}
\begin{figure}[h]
  \centering
    \includegraphics[width=1\columnwidth]{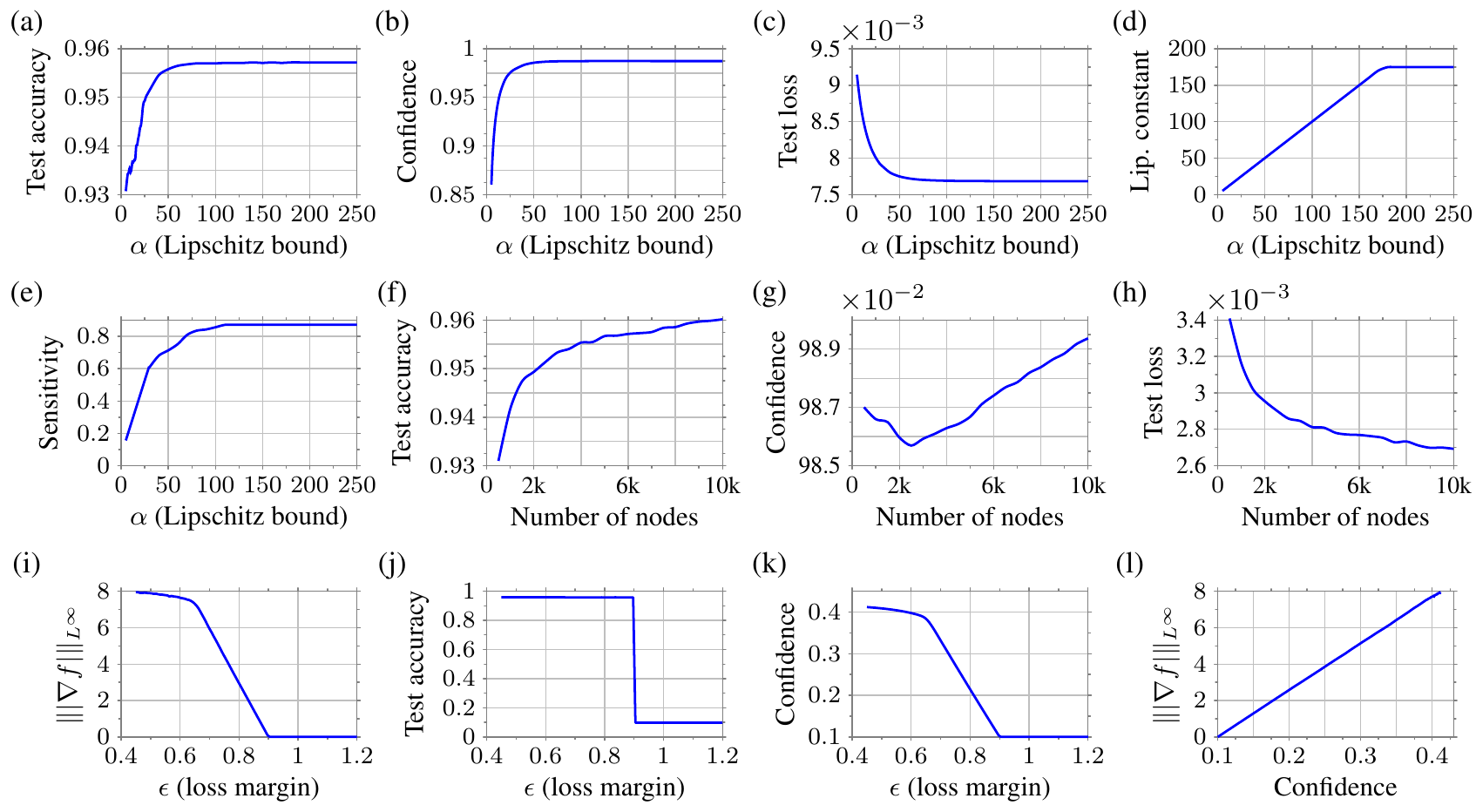} 
    \caption[]{For the standard MNIST dataset,
        panels (a)-(c) show the relationships between Lipschitz
        bound, accuracy, confidence, and loss for Problem
        \eqref{eq:lip_constrained_loss_min}. (d) shows the
        relationship between the Lipschitz constant of the trained model 
        and the Lipschitz bound in \eqref{eq:lip_constrained_loss_min}. 
        We see that the Lipschitz constraint is active for $\alpha<175$, and
        inactive otherwise. (e) shows the confidence degradation under
        bounded perturbation as we vary the Lipschitz bound in
        \eqref{eq:lip_constrained_loss_min}. Panels (f)-(h) show the
        dependence of accuracy, confidence and loss in testing
        on model complexity (number of vertices) for graph-based learning.         
        Panels (i)-(k) show the dependence of the Lipschitz constant,
        accuracy and confidence in testing on the loss margin, 
        for Problem \eqref{eq:lipschitz_min}. Panel (l) shows the
        tradeoff between performance and robustness, seen as an
        increase in the model Lipschitz constant
        with confidence in testing.}
  \label{fig:MNIST} 
\end{figure}

In this section\footnote{The code from numerical experiments in this paper 
is available on GitHub: \text{https://github.com/abedmakdah/Lipschitz-Bounds-and-Provably-Robust-Training-by-Laplacian-Smoothing.git}}, we present the results from numerical experiments
on the standard MNIST dataset of handwritten digits~\cite{YL-CC-CJCB:98},
for the training schemes in Sections \ref{sec:
  robust_min} and \ref{sec: bounds}. We first obtain $n=5000$
graph vertices using the K-means algorithm on the images in the MNIST dataset. 
We then construct a graph $\mathcal{G}=(\mathcal{V},\mathcal{E})$ by
connecting each vertex to its $5$ nearest neighbors, and compute the
solution $\mathbf{v}^*$ to \eqref{eq:p-d_Lipschitz_const_loss_min} for
different values of the Lipschitz bound $\alpha$. We associate each
testing data sample with the closest vertex, evaluate the classification
confidence of $\mathbf{v}^*$, and assign to it the class that
corresponds to the largest confidence. Fig. \ref{fig:MNIST}(a)-(c)
show the dependence of testing accuracy, testing confidence, and testing loss 
on the Lipschitz bound $\alpha$. It can be seen
that both accuracy and confidence increase with the Lipschitz bound,
while the testing loss decreases with the Lipschitz bound. Fig
\ref{fig:MNIST}(d) shows the relationship between the classifier's
Lipschitz constant and the Lipschitz bound $\alpha$. It can be seen
that the constraint in \eqref{eq:lip_constrained_loss_min} is active
for $\alpha<175$, and inactive otherwise. Fig. \ref{fig:MNIST}(e) 
shows the dependence of the classifier's sensitivity to bounded perturbations,
on the Lipschitz bound $\alpha$. 
The sensitivity of the trained classifier is the norm of the 
difference between the nominal and the perturbed confidence 
(confidence degradation). We observe that the sensitivity increases with the Lipschitz bound.
Next, we fix the
Lipschitz bound at $\alpha=300$ and vary the complexity of the
classifier by changing the number of vertices. We observe in Fig.
\ref{fig:MNIST}(f)-(h) that the testing accuracy increases and the
loss decreases with the number of vertices (model complexity), 
while the confidence remains almost constant.
Finally, we fix the number of vertices at $n=5000$ and compute the
solution $\mathbf{v}^*$ to \eqref{eq:W^1,p_min_graph_discretization}
for different values of the loss margin $\epsilon$. Fig.
\ref{fig:MNIST}(i)-(k) show the dependence of the 
Lipschitz constant, testing accuracy and confidence on the
loss margin. As predicted by our theory,
and in accordance with the results obtained in the other numerical
examples in Fig. \ref{fig:MNIST}(a), the classifier's Lipschitz
constant (Fig. \ref{fig:MNIST}(i)), accuracy and confidence (Fig.
\ref{fig:MNIST}(j),(k)) are decreasing functions of the classifier's loss
margin. On the other hand, the model Lipschitz constant is
directly proportional to the classification confidence (Fig. \ref{fig:MNIST}(l)). 
This confirms the existence of a tradeoff between robustness and
performance, and provides a limiting benchmark for comparison with other
models.


\section{Conclusion}\label{sec: conclusion}
In this paper we propose a novel framework to train models with
provable robustness guarantees. At its core, our framework relies on
formulating a provably robust learning problem as a (convex) Lipschitz
constrained loss minimization problem, for which we characterize and
compute the solution by graph-based discretization and discrete heat
flows. Our analysis defines a link between the properties of elliptic
operators and adversarial learning, which provides us with a new
perspective and powerful tools to investigate robustness properties of
the minimizers. Following a similar analysis, we also study the
complementary problem of improving the robustness of a model under a
margin on the loss.  We show that the two notions are tightly related,
and that improving robustness necessarily leads to the deterioration
of the performance of the model (in typical regimes). This
robustification problem, which can be solved using an iterative
procedure based on discrete heat flows involving the $p$-Laplacian,
leads to the characterization of a fundamental tradeoff between the
robustness of a model and its loss, thereby extending and generalizing
recent results relating robustness and performance in adversarial
machine learning. We illustrate our results via academic and a
standard benchmark.

The ideas presented in this paper are of broad interest to the machine
learning community and potentially open up a number of research
directions.  For instance, quantifying the optimality gap of
minimizers of~\eqref{eq:empirical_loss_min_graph_discretization}
with respect to the minimizer for Problem~\eqref{eq:lip_constrained_loss_min}, 
for finite values of~$n$ and~$N$, 
under different Lipschitz bounds, interpolation schemes, and
graph structures, 
will shed light on the underlying fundamental 
relationships between model complexity,
generalization performance, and robustness in graph-based learning.


\newpage

\section{Broader impact}
This paper is primarily of a theoretical nature. We expect our findings to  
impact the development of a formal theory of adversarially robust learning. 
Furthermore, we expect the proposed robust training schemes to 
contribute to efforts in adversarially robust graph-based learning.
However, we do not envision any immediate
application of our results to a societally relevant problem.

\section{Funding disclosure}
This work was supported in part by awards ARO-71603NSYIP,
ONR-N00014-19-1-2264, and AFOSR-FA9550-20-1-0140.

\bibliographystyle{unsrt}
\bibliography{alias,Main,FP,New}

\newpage
\begin{appendices}
\renewcommand\thefigure{\thesection.\arabic{figure}}    
\section{Mathematical preliminaries}
We introduce some mathematical preliminaries related to function spaces
useful in developing our results. 
In what follows, we let~$\mathbb{X} \subset \real^{\dim(\mathbb{X})}$ 
and~$\mathbb{Y} \subset \real^{\dim(\mathbb{Y})}$ be compact
and convex.

\textbf{$L^p$ and $W^{1,p}$ spaces.}
The space~$L^p(\mathbb{X}, \mu)$ of~$p$-integrable functions on~$\mathbb{X}$ 
with respect to an underlying (absolutely continuous) probability 
measure~$\mu \in \mathcal{P}(\mathbb{X})$, is defined as:
\begin{align*}
	L^p(\mathbb{X}, \mu) = \left \lbrace f: \mathbb{X} \rightarrow \real \; \left | \; f ~\text{measurable}~,~ \int_{\mathbb{X}} |f|^p d\mu < \infty \right. \right \rbrace.
\end{align*}
The Sobolev space~$W^{1,p}(\mathbb{X}, \mu)$ is defined as:
\begin{align*}
	W^{1,p}(\mathbb{X}, \mu) = \left \lbrace f \in L^p(\mathbb{X}, \mu) \; \left | \;  \int_{\mathbb{X}} |\nabla f|^p d\mu < \infty \right. \right \rbrace.
\end{align*}
For~$p = \infty$ in the above definitions, we get the space~$L^{\infty}(\mathbb{X}, \mu)$ of 
essentially bounded measurable functions on~$(\mathbb{X}, \mu)$ and
the space~$W^{1,\infty}(\mathbb{X}, \mu)$ of essentially bounded measurable functions 
with essentially bounded measurable gradients on~$(\mathbb{X}, \mu)$.

Now, for~$1 \leq p \leq \infty$,~$L^p((\mathbb{X}, \mu); \mathbb{Y})$ is the space of measurable maps
from~$\mathbb{X}$ to~$\mathbb{Y}$ such that~$|f| \in L^p(\mathbb{X}, \mu)$ for 
any~$f \in L^p((\mathbb{X}, \mu); \mathbb{Y})$, where~$| \cdot |$ is the H-S norm in~$\mathbb{Y}$.
Moreover,~$W^{1,p}((\mathbb{X}, \mu); \mathbb{Y})$ is the space of measurable maps
such that~$|f| \in L^p(\mathbb{X}, \mu)$ and~$|\nabla f| \in L^p(\mathbb{X}, \mu)$
for any~$f \in W^{1,p}((\mathbb{X}, \mu); \mathbb{Y})$.

\textbf{Lipschitz-continuous maps.}
The space~$\Lip(\mathbb{X}; \mathbb{Y})$  
of Lipschitz-continuous maps from~$\mathbb{X}$
to~$\mathbb{Y}$ is such that for any $f \in \Lip(\mathbb{X}; \mathbb{Y})$, 
we have $\left| f(x_1) - f(x_2) \right| \leq \lip(f)
\left| x_1 - x_2 \right|$, where $\lip(f)$ is the Lipschitz constant
of $f$.
From Rademacher's theorem~\cite{LCE:98},
every~$f \in \Lip(\mathbb{X}; \mathbb{Y})$ is almost everywhere
differentiable in~$\mathbb{X}$ (with (a.e.)  gradient~$\nabla f$,
which is also its weak gradient). Further,~$\| |\nabla f| \|_{L^{\infty}(\mathbb{X})} = \lip(f)$ and
we get~$\Lip(\mathbb{X}; \mathbb{Y}) =W^{1,\infty}(\mathbb{X};\mathbb{Y})$.
\section{Robustness to adversarial perturbations and the Lipschitz constant}
In this section, we establish the dependence of sensitivity to adversarial 
perturbations of the loss on the Lipschitz constant
of the input-output map.
Recall from~\eqref{eq:adv_robust_expected_loss_min} that the loss~$L_{\sigma}$ is given by:
\begin{align*}
	L_{\sigma}(f) = \mathbb{E}_{(x,y) \sim \sigma} \left[ \ell(f(x),y) \right].
\end{align*}
Adversarial perturbations~\cite{CS-WZ-IS-JB-DE-IG-RF:14} 
are a subset of perturbations on the data-generating distribution~$\sigma$
generated by bounded maps~$T$ that perturb the
inputs~$x \in \mathbb{X}$ while preserving the
outputs~$y \in \mathbb{Y}$.
We illustrate this for a classification problem:
Let~$(x,y)$ be a true input-label pair in the (nominal) dataset
and~$f$ be a classifier that locally assigns to an input~$x \in \mathbb{X}$
the label~$f(x) \in \mathbb{Y}$. Let~$r$ 
be a minimal perturbation on the input~$x$, given a target label~$y' \in \mathbb{Y}$,
such that~$f(x+r) = y'$ (where~$y'$ is typically chosen to be an incorrect label for~$x$,
that is, $y' \neq y$). 
Now, an adversarial perturbation for the classifier~$f$ 
is generated by the replacement of~$(x,y)$
by~$(x+r, y)$ in the dataset.
To formalize this, we define the class of maps:
\begin{align*}
  \mathcal{T} = \left \lbrace T 
  \; \left| \; T(x,y) = (T_1(x,y) \; , \; y),~ \text{and}~~T_1(x,y)
  \in B_{\delta}(x) \cap \mathbb{X} \right. \right \rbrace,
\end{align*}
where~$B_{\delta}(x)$ is the open ball in~$\real^{\dim(\mathbb{X})}$
of radius~$\delta > 0$ and centered at~$x$. 
Now, adversarial perturbations on the data-generating distribution~$\sigma$
are a subset of perturbations generated by the class~$\mathcal{T}$.

We first characterize the bound on the perturbation of the loss
due to perturbations on~$\sigma$ generated by the class~$\mathcal{T}$.
The perturbation by~$T \in \mathcal{T}$ of the probability measure~$\sigma$
yields the perturbed probability measure~$T_{\#} \sigma$, where~$T_{\#}\sigma$ is the pushforward of~$\sigma$ by the
  map~$T$\footnote{Given a measurable map~$T: \mathbb{Z} \rightarrow \mathbb{Z}'$
and a probability measure~$\sigma \in \mathcal{P}(\mathbb{Z})$, we let~$T_{\#} \sigma$
denote the pushforward of~$\sigma$ by the map~$T$, 
where for any Borel measurable set~$B \subset \mathbb{Z}'$ we have~$T_{\#} \sigma (B) = \sigma (T^{-1}(B))$.}.
We note that the perturbation of the loss~$\left| L_{T_{\#}\sigma}(f) - L_{\sigma}(f) \right|$ satisfies:
\begin{align*}
	\left| L_{T_{\#}\sigma}(f) - L_{\sigma}(f) \right| &= \left| \mathbb{E}_{(x,y) \sim T_{\#}\sigma} \left[ \ell(f(x), y) \right] 
																										-	\mathbb{E}_{(x,y) \sim \sigma} \left[ \ell(f(x), y) \right] \right| \\
																					&= \left| \int_{\mathbb{X} \times \mathbb{Y}} \ell(f(x), y) d \left(T_{\#}\sigma \right)(x,y) 
																										- \int_{\mathbb{X} \times \mathbb{Y}} \ell(f(x), y) d\sigma (x,y)  \right|  \\
																					&= \left| \int_{\mathbb{X} \times \mathbb{Y}} \left( \ell(f(T_1(x,y)), y)
																										- \ell(f(x), y) \right) d\sigma (x,y)  \right| \\
																					&\leq \lip(\ell) \lip(f) \left| \int_{\mathbb{X}} \left(T_1(x,y) - x \right) d\mu(x)  \right| \\
																					&\leq \lip(\ell) \lip(f) \delta.
\end{align*} 

We next characterize the sensitivity of the loss for a given~$f$ 
to perturbations on the data-generating distribution
generated by the class~$\mathcal{T}$.
Let a family of transport maps~$T^h = (1-h) \id + h T$ for some~$T \in \mathcal{T}$
and $h \in [0,1]$ (with $\id$ being the identity map), perturb the data-generating
distribution~$\sigma$ as $\sigma^h = T^h_{\#} \sigma$.
The (Gateaux) derivative of the loss along the family of adversarial perturbations~$T^h$, is now given by:
\begin{align*}
	D^{(T)} L_{\sigma}(f) = \left. \frac{d}{dh} L_{\sigma^h}(f) \right|_{h = 0} &= \lim_{h \rightarrow 0} \frac{L_{\sigma^h}(f) - L_{\sigma}(f)}{h} \\
							&= \lim_{h \rightarrow 0} \frac{1}{h} \int_{\mathbb{X} \times \mathbb{Y}} \left[ \ell(f(T^h(x,y)), y) - \ell(f(x), y) \right] d\sigma(x,y).							
\end{align*}
We note that $\left| \frac{\ell(f(T^h(x,y)),y) - \ell(f(x),y)}{h} \right| \leq \lip(\ell) \left| \frac{f(T^h(x,y)) - f(x)}{h} \right|
\leq \lip(\ell) \lip(f)  \frac{\left| T^h(x,y) - x \right|}{h} = \lip(\ell) \lip(f) \left| T_1(x,y) - x \right|$.
It then follows from the Dominated Convergence Theorem~\cite{WR:64} that:
\begin{align*}
		D^{(T)} L_{\sigma}(f)	&= \int_{\mathbb{X} \times \mathbb{Y}} \left \langle \nabla_1\ell (f(x), y) \cdot \nabla f(x) \; , \; T_1(x,y) - x \right \rangle d\sigma(x,y) \\
							&= \mathbb{E}_{(x,y) \sim \sigma} \left[ \left \langle \nabla_1\ell (f(x), y) \cdot \nabla f(x) \; , \; T_1(x,y) - x \right \rangle \right].
\end{align*}
We now define the sensitivity as the worst-case increase of the loss functional following an
adversarial perturbation. That is, the sensitivity of the loss 
is the~$L^{\infty}$-norm (with respect to the measure~$\sigma$) of the
gradient~$\nabla_1 \ell \cdot \nabla f$ 
(precisely, $\| | \nabla_1 \ell \cdot \nabla f | \|_{L^{\infty}(\mathbb{X} \times \mathbb{Y}, \sigma)}$),
which satisfies the bound:
\begin{align*}
  \underbrace{\| | \nabla_1 \ell \cdot \nabla f | \|_{L^{\infty}(\mathbb{X} \times
  \mathbb{Y}, \sigma)}}_{\text{sensitivity of $L$ to adv. perturbation}} \leq \underbrace{\| | \nabla_1 \ell |
  \|_{L^{\infty}(\mathbb{X} \times \mathbb{Y},
  \sigma)}}_{\text{Lipschitz constant of $\ell$}}  \cdot
  \underbrace{\| | \nabla f | \|_{L^{\infty}(\mathbb{X},
  \mu)}}_{\text{Lipschitz constant of $f$}}
\end{align*}
where $\mu$ is the marginal of~$\sigma$ over~$\mathbb{X}$, and
$\| | \nabla f | \|_{L^{\infty}(\mathbb{X}, \mu)}$ is the Lipschitz
constant of~$f$ over the support of~$\mu$.

We therefore get that the sensitivity
of the loss functional to adversarial perturbations is indeed
modulated by the Lipschitz constant of the input-output mapping. 
Thus, restricting the search space to the class of Lipschitz maps with a
bound~$\alpha \geq 0$ on the Lipschitz constant, as in the
minimization problem~\eqref{eq:adv_robust_expected_loss_min}, is
convenient for analysis, and does not restrict the generality of the
adversarially robust learning problem, and it allows us to obtain 
adversarially robust minimizers of the loss~$L_{\sigma}$.
\section{The Lipschitz-constrained loss minimization problem~\eqref{eq:adv_robust_expected_loss_min} is convex}
We recall that Problem~\eqref{eq:adv_robust_expected_loss_min} is given by:
\begin{align*}
  \inf_{f \in \Lip(\mathbb{X}, \mu) }~ \left \lbrace
  \underbrace{\mathbb{E}_{(x,y) \sim \sigma} \left[ \ell
  \left( f(x), y \right) \right]}_{\triangleq L_{\sigma}(f)}
  \qquad  \text{s.t.}~~\lip(f) \leq \alpha \right \rbrace ,
\end{align*}
where $\sigma$ is an absolutely continuous probability measure
on~$\mathbb{X} \times \mathbb{Y}$ and the loss
function~$\ell: \mathbb{Y} \times \mathbb{Y} \rightarrow
\realnonnegative$ is strictly convex and Lipschitz continuous
and~$\alpha \geq 0$.

Firstly, we get that the loss~$L_{\sigma}$ in~\eqref{eq:adv_robust_expected_loss_min} is strictly
convex. 
To see this, let~$f_1, f_2 \in \Lip(\mathbb{X}, \mu)$ be such that~$L_{\sigma}( f_1) < \infty$ 
and~$L_{\sigma}( f_2) < \infty$.
For~$t \in [0,1]$, we get from the convexity of~$\Lip(\mathbb{X}, \mu)$ 
that~$t f_1 + (1-t) f_2 \in \Lip(\mathbb{X}, \mu)$. Also, from the strict convexity of the
loss function~$\ell$, we get:
\begin{align*}
	L_{\sigma}( t f_1 + (1-t) f_2 ) &= \mathbb{E}_{(x,y) \sim \sigma} \left[ \ell( (t f_1 + (1-t) f_2)(x) , y) \right] \\
													&= \mathbb{E}_{(x,y) \sim \sigma} \left[ \ell( t f_1(x) + (1-t) f_2(x) , y) \right] \\
													&\leq \mathbb{E}_{(x,y) \sim \sigma} \left[  t \ell( f_1(x), y) + (1-t) \ell(f_2(x) , y) \right] \\
													&= t\mathbb{E}_{(x,y) \sim \sigma} \left[  \ell( f_1(x), y) \right] + (1-t) \mathbb{E}_{(x,y) \sim \sigma} \left[ \ell(f_2(x) , y) \right]  \\
													&= t L_{\sigma}(f_1) + (1-t) L_{\sigma}(f_2).
\end{align*}
Moreover, the inequality is strict for~$t \in (0,1)$, from which it follows
that the loss~$L_{\sigma}$ is strictly convex.

Now, let~$f_1, f_2 \in \Lip(\mathbb{X}, \mu)$ such that~$\lip(f_1) \leq \alpha$
and~$\lip(f_2) \leq \alpha$. For the map~$\lambda f_1 + (1-\lambda) f_2$,
$\lambda \in [0,1]$, and~$x_1, x_2 \in \mathbb{X}$, it follows that:
\begin{align*}
	| (\lambda f_1 + (1-\lambda) f_2)(x_1) &- (\lambda f_1 + (1-\lambda) f_2)(x_2) |  \\
					&= \left| \lambda \left( f_1(x_1) - f_1(x_2) \right) + (1-\lambda) (f_2(x_1) - f_2(x_2))  \right| \\
					&\leq \lambda \left| f_1(x_1) - f_1(x_2) \right| + (1-\lambda) \left| f_2(x_1) - f_2(x_2) \right| \\
					&\leq \lambda	\lip(f_1) \left| x_1 - x_2 \right|	+ (1-\lambda) 	\lip(f_2) \left| x_1 - x_2 \right| \\
					&\leq \alpha \left| x_1 - x_2 \right|,
\end{align*}
and we get~$\lip(\lambda f_1 + (1-\lambda) f_2) \leq \alpha$. 
Therefore, the constraint in~\eqref{eq:adv_robust_expected_loss_min}
is convex.
From strict convexity of the loss~$L_{\sigma}$ and convexity of 
the constraint set~$\left \lbrace f \in \Lip((\mathbb{X}, \mu), \mathbb{Y})
\; | \; \lip(f) \leq \alpha \right \rbrace$, we get that Problem~\eqref{eq:adv_robust_expected_loss_min}
is convex.
\section{Proof of Theorem~\ref{thm:saddle_point_Lip_constraint} (Saddle point of Lagrangian~$\mathcal{L}$)}
\label{sec:saddle_point_proof_Sec_2}
%
%
\emph{(i) Derivative of loss function~$L$ w.r.t~$f$.}
We have:
\begin{align*}
	L_{\sigma}(f) = \mathbb{E}_{x \sim \mu} \left[ \mathbb{E}_{y \sim \pi(y \; | \; x)}  \left[ \ell(f(x),y) \right] \right],
\end{align*}
where~$\mu$ is the marginal over~$\mathbb{X}$
and $\pi$ the conditional of the joint 
distribution~$\sigma \in \mathcal{P}(\mathbb{X} \times \mathbb{Y})$.
Let~$\left \lbrace f^{\epsilon} \right \rbrace_{\epsilon \in [0,	1]}$
be a family of maps from~$\mathbb{X}$ to~$\mathbb{Y}$
that is pointwise smooth (i.e., for any~$x \in \mathbb{X}$,
$F(\epsilon, x) = f^{\epsilon}(x)$ is smooth in~$\epsilon$).
We now evaluate the derivative of the loss function
$L_{\sigma}$ w.r.t. the family~$\left \lbrace f^{\epsilon} \right \rbrace_{\epsilon \in [0,1]}$,
at~$\epsilon = 0$, as follows:
\begin{align*}
	\frac{d L_{\sigma}}{d\epsilon} (f^0) &= \lim_{\epsilon \rightarrow 0} \frac{L_{\sigma}( f^{\epsilon}) - L_{\sigma}(f^0)}{\epsilon} \\
				&= \lim_{\epsilon \rightarrow 0} \frac{1}{\epsilon} \int_{\mathbb{X}} \left[ \int_{\mathbb{Y}} \left( \ell(f^{\epsilon}(x),y) - \ell(f^0(x),y) \right) d\pi(y \; | \; x) \right] d\mu(x).
\end{align*}
We note that $\left| \frac{\ell(f^{\epsilon}(x),y) - \ell(f^0(x),y)}{\epsilon} \right| \leq \lip(\ell) \left| \frac{f^{\epsilon}(x) - f^0(x)}{\epsilon} \right|
\leq \lip(\ell) \lip(F(\cdot, x))$, where~$\lip(F(\cdot, x))$ is the Lipschitz constant of
$F$ as a function of~$\epsilon$ at every~$x \in \mathbb{X}$ (since~$F(\cdot, x)$
is smooth in~$[0,1]$ for every~$x \in \mathbb{X}$, it is also Lipschitz continuous).
It then follows from the Dominated Convergence Theorem~\cite{WR:64} that:
\begin{align*}
	\frac{d L_{\sigma}}{d\epsilon} (f^0) &= \lim_{\epsilon \rightarrow 0} \frac{1}{\epsilon} \int_{\mathbb{X}} \left[ \int_{\mathbb{Y}} \left( \ell(f^{\epsilon}(x),y) - \ell(f^0(x),y) \right) d\pi(y \; | \; x) \right] d\mu(x) \\
	&= \int_{\mathbb{X}} \left[ \int_{\mathbb{Y}} \lim_{\epsilon \rightarrow 0} \frac{1}{\epsilon} \left( \ell(f^{\epsilon}(x),y) - \ell(f^0(x),y) \right) d\pi(y \; | \; x) \right] d\mu(x) \\
	&= \int_{\mathbb{X}} \left[ \int_{\mathbb{Y}} \nabla_1\ell (f^0(x),y) \cdot \left. \frac{\partial f^{\epsilon}}{\partial \epsilon}(x) \right|_{\epsilon = 0} d\pi(y \; | \; x) \right] d\mu(x) \\
	&=  \int_{\mathbb{X}} \left[ \int_{\mathbb{Y}} \nabla_1 \ell(f^0(x),y) ~d\pi(y \; | \; x) \right]\cdot \left. \frac{\partial f^{\epsilon}}{\partial \epsilon} (x) \right|_{\epsilon = 0} d\mu(x) \\
	&= \int_{\mathbb{X}} \frac{\partial \bar{L}}{\partial f} \cdot \left. \frac{\partial f^{\epsilon}}{\partial \epsilon}\right|_{\epsilon = 0} d\mu(x),
\end{align*}
where we denote by $\partial_f \bar{L}_{\sigma} = \frac{\partial \bar{L}_{\sigma}}{\partial f} = \int_{\mathbb{Y}} \nabla_1 \ell(f^0(x),y) ~d\pi(y \; | \; x)$ the
functional derivative of~$\bar{L}_{\sigma}$ w.r.t.~$f$.

\emph{(ii) Minimizer of~\eqref{eq:lip_constrained_loss_min}.}
The search space for Problem~\eqref{eq:lip_constrained_loss_min} is given by,
\begin{align*}
	\mathcal{F} = \left \lbrace f \in W^{1,\infty}((\mathbb{X}, \mu), \mathbb{Y}) \; | \; \| |\nabla f| \|_{L^{\infty}(\mathbb{X}, \mu)} \leq \alpha \right \rbrace.
\end{align*}
We see that~$\mathcal{F}$ is closed, convex and bounded. Boundedness of~$\mathcal{F}$ follows
from compactness of~$\mathbb{Y}$ which implies that there exists an~$M \in \realnonnegative$ such that
$\mathbb{Y} \subset B_M(\mathbf{0}_{\mathbb{Y}})$. It follows that for any~$f \in \mathcal{F}$,
we have~$\| |f| \|_{L^{\infty}(\mathbb{X}, \mu)} \leq M$. Moreover, we have~$\| |\nabla f| \|_{L^{\infty}(\mathbb{X}, \mu)} \leq \alpha$.
Therefore, $\| f \|_{W^{1,\infty}((\mathbb{X}, \mu), \mathbb{Y})} 
= \| |f| \|_{L^{\infty}(\mathbb{X}, \mu)} + \| |\nabla f| \|_{L^{\infty}(\mathbb{X}, \mu)} \leq M + \alpha < \infty$
for any~$f \in \mathcal{F}$.

The loss~$L_{\sigma}$ is strictly convex and lower semicontinuous (in fact, it is (Gateaux) differentiable as seen earlier
for absolutely continuous~$\sigma$, since~$\ell$ is strictly convex and Lipschitz-continuous).

Let~$\lbrace f_n \rbrace_{n \in \mathbb{N}}$ be a minimizing sequence in~$\mathcal{F}$ for
the loss~$L_{\sigma}$, such that~$f_n \in \mathcal{F}$
and~$\lim_{n \rightarrow \infty} L_{\sigma}(f_n)
= \inf_{f \in \mathcal{F}} L_{\sigma}(f)$. Clearly,
the sequence~$\lbrace f_n \rbrace_{n \in \mathbb{N}}$ is uniformly
bounded since~$\| f_n \|_{W^{1,\infty}((\mathbb{X}, \mu), \mathbb{Y})} \leq M + \alpha$.
It is also uniformly equicontinuous, since~$\left| f_n(x_1) - f_n(x_2) \right| \leq \alpha | x_1 - x_2 |$
for all~$n \in \mathbb{N}$. Therefore, by the Arzel{\`a}-Ascoli Theorem~\cite{WR:64}, 
there exists a uniformly converging subsequence~$\lbrace f_{n_j} \rbrace_{j \in \mathbb{N}}$,
with the limit~$f^* \in \mathcal{F}$. Furthermore, by the continuity of~$L_{\sigma}$,
we get~$\lim_{j \rightarrow \infty} L_{\sigma}(f_{n_j}) = L_{\sigma}(f^*) 
= \min_{f \in \mathcal{F}} L_{\sigma}(f)$. By the strict convexity of the loss~$L_{\sigma}$,
we get that~$f^*$ is the unique global minimizer of~$L_{\sigma}$.

Thus, Problem~\eqref{eq:lip_constrained_loss_min} has a unique global 
minimizer~$f^* \in \left \lbrace f \in W^{1,\infty}((\mathbb{X}, \mu), \mathbb{Y}) \; | \;
\lip(f) \leq \alpha \right \rbrace$.

\emph{(iii) Saddle points of Lagrangian functional~$\mathcal{L}_{\sigma}$.}
The constraint set is given by~$\lbrace f \in W^{1,\infty}((\mathbb{X}, \mu), \mathbb{Y})
\; | \;  \mathcal{G}(f) \in ( -\infty, 0]  \rbrace$, where
$\mathcal{G}(f) = \| G_f \|_{L^{\infty}(\mathbb{X}, \mu)}$,
and we have the constraint qualification:
\begin{align*}
	0 \in  \mathrm{int} \left \lbrace \mathcal{G} \left( W^{1,\infty}((\mathbb{X}, \mu), \mathbb{Y}) \right) + [0, \infty) \right \rbrace,
\end{align*}
where the operation $+$ denotes the Minkowski sum. 
This allows us to apply Theorem~3.6 in \cite{JFB-AS:13}
to infer that the set of Lagrange multipliers corresponding to the (unique) minimizer~$f^*$
is a non-empty, convex, bounded and weakly$-^*$ compact subset of
$L^{\infty}(\mathbb{X}, \mu)^*_{\geq 0}$.
Moreover, we note that $(-\infty, 0]$ is a closed convex cone, and it follows from
Theorem~3.4-(iii) in \cite{JFB-AS:13} that for any Lagrange multiplier~$\lambda^*$,
the pair $(f^*, \lambda^*)$ is a saddle point of the Lagrangian
functional~$\mathcal{L}_{\sigma}$. Uniqueness of~$\lambda^*$
again follows from the strict convexity of~$L_{\sigma}$.
We also have the \textit{feasibility} condition~$G_{f^*} \leq 0$
(that is, $\left| \nabla f^* \right| \leq \alpha$) and~$\lambda^* \geq 0$
$\mu$-a.e. in~$\mathbb{X}$.

Now, the (Gateaux) derivative of the Lagrangian $\mathcal{L}_{\sigma}(f, \lambda) = L_{\sigma}(f) + \lambda(G_f)$ 
in $W^{1,\infty}((\mathbb{X}, \mu), \mathbb{Y})$ along $V \in W^{1,\infty}((\mathbb{X}, \mu), \mathbb{Y})$
is given by:
\begin{align*}
	D_1^{(V)} \mathcal{L}_{\sigma}(f, \lambda) = \int_{\mathbb{X}} \partial_f \bar{L}_{\sigma} \cdot V ~d\mu 
																					+ \int_{\mathbb{X}} \nabla f \cdot \nabla V~d(\lambda \mu),
\end{align*}
where~$D_1^{(V)}$ denotes the directional derivative of the first argument along~$V$
and~$\lambda \mu$ is an absolutely continuous measure ($\lambda$-weighting on
the underlying measure~$\mu$. Recall that~$\lambda \in L^{\infty}(\mathbb{X}, \mu)^*_{\geq 0}$
is itself a bounded, finitely additive absolutely continuous measure).
The above expression can be derived using a similar construction of a limit
and the application of the Dominated Convergence Theorem as earlier in this section.
%
%
  
 By the Minimax Theorem, we have~$\mathcal{L}_{\sigma}(f^*, \lambda^*)
 = \inf_{f} \sup_{\lambda} \mathcal{L}_{\sigma}(f, \lambda) 
 = \sup_{\lambda} \inf_{f} \mathcal{L}_{\sigma}(f, \lambda)$,
 where the infimum is taken over~$W^{1,\infty}((\mathbb{X}, \mu), \mathbb{Y})$
 and the supremum over~$\lambda \in L^{\infty}(\mathbb{X}, \mu)^*_{\geq 0}$.
 We therefore have~$\mathcal{L}_{\sigma}(f^*, \lambda^*) \geq \mathcal{L}_{\sigma}(f^*, 0)$,
 which yields the condition~$\lambda^*(G_{f^*}) \geq 0$. Moreover, from feasibility,
 we have~$G_{f^*} \leq 0$ and~$\lambda^* \geq 0$, which implies that~$\lambda^*(G_{f^*}) \leq 0$.
 This results in the \textit{complementary slackness} condition~$\lambda^*(G_{f^*}) = 0$.
 From the Minimax equality, we get that~$(f^*, \lambda^*)$ is also a 
 critical point of~$\mathcal{L}_{\sigma}$, that is, $D_1^{(V)} \mathcal{L}_{\sigma}(f^*, \lambda^*) = 0$,
 which implies that~$\int_{\mathbb{X}} \partial_f \bar{L}_{\sigma}(f^*) \cdot V ~d\mu 
	+ \int_{\mathbb{X}} \nabla f^* \cdot \nabla V~d(\lambda^* \mu) = 0$, 
	which is the \textit{stationarity} condition.
%
%

{\emph{(iv) Improved regularity of Lagrange multiplier~$\lambda^*$.}}
We can indeed establish stronger regularity for the Lagrange multiplier~$\lambda^*$.
We have that the Lagrange multipliers $\lambda^* \in L^{\infty}(\mathbb{X}, \mu)^*_{\geq 0}$,
which is a bounded, finitely additive measure absolutely continuous measure,
is also a linear continuous functional on~$L^{\infty}(\mathbb{X}, \mu)$ and must therefore
vanish on sets of $\mu$-measure zero (i.e., $\lambda^*(A) = 0$ for
$A \subset \mathbb{X}$ with $\mu(A) = 0$).
Moreover, from Theorem~1.24 in~\cite{KY-EH:52},
we can decompose $\lambda^* = \lambda^*_c + \lambda^*_p$,
where $\lambda^*_c$ is a non-negative countably additive measure
and $\lambda^*_p$ is non-negative and purely finitely additive.
By the Radon-Nikodym theorem, we get that there exists 
a function~$h_c \in L^1(\mathbb{X}, \mu)$ such that the countably additive
and absolutely continuous measure~$\lambda^*_c$ 
satisfies $d\lambda^*_c = h_c~d\mu$.
By substitution in the stationarity condition,
we get~$\int_{\mathbb{X}} \partial_f \bar{L}_{\sigma} \cdot V d\mu = 
- \int_{\mathbb{X}} \nabla f^* \cdot \nabla V~d(\lambda^*_c \mu)
- \int_{\mathbb{X}} \nabla f^* \cdot \nabla V~d(\lambda^*_p \mu)$.
We now consider a set $D_{\delta} = \left \lbrace x \in \mathbb{X} \; | \;
 -\delta \leq G_{f^*}(x) \leq 0  \right \rbrace$,
 with $0 < \delta < \alpha^2$.
 By complementary slackness, we note that $\lambda^*(\mathbb{X} 
 \setminus D_{\delta}) = 0$. Since $\lambda^*_p$ is
 purely finitely additive, it implies that there must exist 
 a collection of nonempty sets $\lbrace E_n \rbrace_{n \in \mathbb{N}}$ with
 $E_{n+1} \subset E_n$ and $\lim_{n \rightarrow \infty} E_n = \emptyset$,
  such that $\lim_{n \rightarrow \infty} \lambda^*_p(E_n) > 0$\footnote{
  For a countably additive measure~$\nu$ that is absolutely continuous w.r.t. the
  Lebesgue measure, and
  any collection of nonempty sets $\lbrace E_n \rbrace_{N \in \mathbb{N}}$ with
  $E_{n+1} \subset E_n$ and $\lim_{n \rightarrow \infty} E_n = \emptyset$,
  we have $\lim_{n \rightarrow \infty} \nu(E_n) = 0$~\cite{KY-EH:52}.}.
  Since $\lambda^*(\mathbb{X}  \setminus D_{\delta}) = 0$, we can suppose
  without loss of generality that $E_0 \subset D_{\delta}$.
 We also consider another collection of nonempty sets $\lbrace E'_n \rbrace_{n \in \mathbb{N}}$,
 with the same properties (with $E'_0 \subset D_{\delta}$,
  $E'_{n+1} \subset E'_n$ and $\lim_{n \rightarrow \infty} E'_n = \emptyset$), 
 such that $E_n \subset E'_n$ for all $n \in \mathbb{N}$. 
 We note that for $x \in D_{\delta}$, we have $0 < \alpha^2 - \delta \leq |\nabla f^*(x)|^2 \leq \alpha^2$,
 which implies that $\nabla f^*$ does not vanish on~$E'_n$ for any $n \in \mathbb{N}$.
 We now consider a family of variations $V_n \in W^{1,\infty}(\mathbb{X},\mu)$ for $n \in \mathbb{N}$
 such that~$V_n$ and~$\nabla V_n$ are supported in~$E'_n$,
 $\nabla f^* \cdot \nabla V_n \geq 0$ in~$E'_n$
 and $\nabla f^* \cdot \nabla V_n \geq \epsilon$ in~$E_n$ (uniformly).
 The stationarity condition now yields, for~$n \in \mathbb{N}$:
 \begin{align*}
 	- \int_{E'_n} \partial_f \bar{L}_{\sigma}(f^*) \cdot V_n  d\mu &=  \int_{E'_n} (\nabla
     f^* \cdot \nabla V_n)~h_c d\mu + \int_{E'_n} \nabla
     f^* \cdot \nabla V_n~ d(\lambda^*_p \mu) \\
     &\geq  \int_{E'_n} (\nabla
     f^* \cdot \nabla V_n)~h_c d\mu + \epsilon \int_{E_n} d(\lambda^*_p\mu).
 \end{align*}
 In the limit $n \rightarrow 0$, we have 
 $\lim_{n \rightarrow \infty} \int_{E'_n} \partial_f \bar{L}_{\sigma}(f^*) \cdot V_n  d\mu = 0$
 and $\lim_{n \rightarrow \infty} \int_{E'_n} (\nabla
     f^* \cdot \nabla V_n)~h_c d\mu = 0$, which 
     implies that $0 \leq  \lim_{n \rightarrow \infty} \epsilon \int_{E_n} d(\lambda^*_p \mu)
     \leq 0$, and we get $\lim_{n \rightarrow \infty} \lambda^*_p (E_n) = 0$, i.e., 
     the measure~$\lambda^*$ does not have a purely finitely additive component.
     Therefore, the measure~$\lambda^*$ is countably additive (and absolutely continuous) 
     and possesses a Radon-Nikodym derivative w.r.t.~$\mu$, in~$L^1(\mathbb{X}, \mu)$. 
     For ease of notation, we henceforth let $\lambda^* \in L^1(\mathbb{X}, \mu)$ 
     also denote its density function.

  Since $\lambda^* \in L^1(\mathbb{X}, \mu)_{\geq 0}$
  and $G_{f^*} \leq 0$ $\mu$-a.e. in~$\mathbb{X}$,
  we can now indeed state the complementary slackness condition as 
  $\lambda^* \left( |\nabla f^*| - \alpha \right) = 0$ $\mu$-a.e. in~$\mathbb{X}$.

 Moreover, the stationarity condition, under~$\lambda^* \in L^1(\mathbb{X}, \mu)_{\geq 0}$
 can now be expressed as:
\begin{align*}
	0 &= \int_{\mathbb{X}} \partial_f \bar{L}_{\sigma}(f^*) \cdot V~d\mu 
		+ \int_{\mathbb{X}} \nabla f^* \cdot \nabla V~\lambda^*~d\mu  \\
		&= \int_{\mathbb{X}} \partial_f \bar{L}_{\sigma}(f^*) \cdot V~d\mu 
		- \int_{\mathbb{X}} \frac{1}{\mu} \nabla \cdot \left( \lambda^* \mu \nabla f^* \right) \cdot V~d\mu
		+ \int_{\partial \mathbb{X}} \lambda^* \nabla f^* \cdot \mathbf{n} V \mu~dS,
\end{align*}
where we have used the Divergence Theorem to obtain the final
equality, with~$S$ as the surface measure on~$\partial \mathbb{X}$. 
As the above holds for any variation~$V \in W^{1,\infty}((\mathbb{X}, \mu),\mathbb{Y})$,
it must follow that~$-\frac{1}{\mu} \nabla \cdot \left( \mu \lambda^* \nabla
  f^* \right) + \partial_f \bar{L}_{\sigma}(f^*) = 0$ $\mu$-a.e. in~$\mathbb{X}$ 
  and~$\lambda^* \mu \nabla f^* \cdot \mathbf{n} = 0$ on~$\partial \mathbb{X}$,
and if we do not suppose stronger regularity of the saddle point~$(f^*, \lambda^*)$,
the equations must be hold weakly.  

The above correspond to the necessary KKT conditions. Conversely, any
solution pair $(f^*, \lambda^*)$ which satisfies the above
KKT conditions is a saddle point for the Lagrangian~$\mathcal{L}_\sigma$ and is
a solution to the original optimization problem. 
%
%
\section{Proof of Theorem~\ref{thm:saddle_point_W^1,p_min} (Saddle points of Lagrangian~$\mathcal{H}$)}
\label{sec:saddle_point_proof_Sec_3}
%
%
%
\emph{(i) Minimizers of~\eqref{eq:W^1,p_min}.}
The search space for Problem~\eqref{eq:W^1,p_min} is given by:
\begin{align*}
	\mathcal{F}_p = \left \lbrace f \in W^{1,p}((\mathbb{X},\mu),\mathbb{Y}) \; | \;
	L_{\sigma}(f) \leq J^*_{\sigma}(\alpha) + \epsilon \right \rbrace.
\end{align*}

Let~$\lbrace u_n \rbrace_{n \in \mathbb{N}}$ be a minimizing
sequence in~$\mathcal{F}_p$ for the~$W^{1,p}$-seminorm, such that
$u_n \in \mathcal{F}_p$ for all~$n \in \mathbb{N}$ and
$\lim_{n \rightarrow \infty} \| |\nabla u_n| \|_{L^p(\mathbb{X}, \mu)} 
= \inf_{u \in \mathcal{F}_p} \| |\nabla u| \|_{L^p(\mathbb{X}, \mu)}$.
Since~$f^* \in W^{1,\infty}((\mathbb{X}, \mu), \mathbb{Y})$, the minimizer
of Problem~\eqref{eq:lip_constrained_loss_min} also belongs to~$\mathcal{F}_p$,
that is, $f^* \in \mathcal{F}_{p}$ and $inf_{u \in \mathcal{F}_p} 
\| |\nabla u| \|_{L^p(\mathbb{X}, \mu)} \leq \| |\nabla f^*| \|_{L^p(\mathbb{X}, \mu)} \leq \alpha$,
we can choose the minimizing sequence to satisfy the bound~$ \| |\nabla u_n| \|_{L^p(\mathbb{X}, \mu)} 
\leq \alpha$. Similar to Section~\ref{sec:saddle_point_proof_Sec_2}, we now
have the uniform bound~$\| u_n \|_{W^{1,p}((\mathbb{X},\mu),\mathbb{Y})}
\leq M + \alpha$ for all~$n \in \mathbb{N}$.
For $p > \dim(\mathbb{X})$, we have from Morrey's Inequality~\cite{LCE:98},
for every~$n \in \mathbb{N}$, that:
\begin{align*}
	\left| u_n(x_1) - u_n(x_2) \right| 
	&\leq \frac{2p \dim(\mathbb{X})}{p - \dim(\mathbb{X})} | x_1 - x_2 |^{1 - \frac{\dim(\mathbb{X})}{p}} \| |\nabla u_n| \|_{L^p(\mathbb{X}, \mu)} \\
	&\leq 2C \dim(\mathbb{X}) (1+\dim(\mathbb{X})) | x_1 - x_2 |^{\frac{1}{1+\dim(\mathbb{X})}} \alpha,
\end{align*}
where~$C = \max \left \lbrace 1, \mathrm{diam}(\mathbb{X})^{\frac{\dim(\mathbb{X})}{1+\dim(\mathbb{X})}} \right \rbrace$.
Thus, the sequence~$\lbrace u_n \rbrace_{n \in \mathbb{N}}$ is also uniformly equicontinuous.
Therefore, by the Arzel{\`a}-Ascoli Theorem, there exists a uniformly converging
subsequence~$\lbrace u_{n_j} \rbrace_{j \in \mathbb{N}}$ with limit~$f^{\epsilon, p} \in \mathcal{F}_p$.
Furthermore, by the continuity of the $W^{1,p}$-seminorm, we get that
$\lim_{j \rightarrow \infty} \| |\nabla u_{n_j}| \|_{L^p(\mathbb{X}, \mu)} 
= \| |\nabla f^{\epsilon, p} | \|_{L^p(\mathbb{X}, \mu)} 
= \min_{f \in \mathcal{F}_p} \| |\nabla f | \|_{L^p(\mathbb{X}, \mu)} $.
By convexity of the $W^{1,p}$-seminorm, we get that~$f^{\epsilon, p}$
is a global minimizer for Problem~\eqref{eq:W^1,p_min}.

We therefore conclude that Problem~\eqref{eq:W^1,p_min} is guaranteed to 
have (atleast one) global 
minimizer~$f^{\epsilon, p} \in  \left \lbrace f \in W^{1,p}((\mathbb{X}, \mu), \mathbb{Y})
\; | \; L_{\sigma}(f) \leq J^*_{\sigma}(\alpha) + \epsilon \right \rbrace$.

\emph{(ii) Saddle points of Lagrangian functional~$\mathcal{H}_{\sigma}$.}
The constraint set is given by~$\lbrace f \in W^{1,p}((\mathbb{X}, \mu), \mathbb{Y})
\; | \; \mathcal{G}(f) \leq 0  \rbrace$, where
$\mathcal{G}(f) =  L_{\sigma}(f) - (J^*_{\sigma}(\alpha) + \epsilon)$,
and we have the constraint qualification:
\begin{align*}
	0 \in  \mathrm{int} \left \lbrace \mathcal{G} \left( W^{1,p}((\mathbb{X}, \mu), \mathbb{Y}) \right) + [0, \infty) \right \rbrace,
\end{align*}
where the operation $+$ denotes the Minkowski sum. 
This allows us to apply Theorem~3.6 in \cite{JFB-AS:13}
to infer that the set of Lagrange multipliers corresponding to the minimizer~$f^{\epsilon, p}$
is a non-empty, convex, bounded and weakly$-^*$ compact subset of
$\realnonnegative$.
Moreover, we note that $(-\infty, 0]$ is a closed convex cone, and it follows from
Theorem~3.4-(iii) in \cite{JFB-AS:13} that for any Lagrange multiplier~$\kappa^{\epsilon, p}$,
the pair $(f^{\epsilon, p}, \kappa^{\epsilon, p})$ is a saddle point of the Lagrangian
functional~$\mathcal{H}_{\sigma}$.
We also have the \textit{feasibility} condition~$L_{\sigma}(f) \leq J^*_{\sigma}(\alpha) + \epsilon$.

Following a similar procedure as in Section~\ref{sec:saddle_point_proof_Sec_2}, we obtain the 
(Gateaux) derivative of the Lagrangian $\mathcal{H}_{\sigma}(f, \kappa) = \frac{1}{p}  \left \|  |\nabla f|  \right \|^p_{L^p(\mathbb{X}, \mu)} 
+ \kappa \left( L_{\sigma}(f) -(J^*_{\sigma}(\alpha) + \epsilon) \right)$ 
in $W^{1,p}((\mathbb{X}, \mu), \mathbb{Y})$ along $V \in W^{1,p}((\mathbb{X}, \mu), \mathbb{Y})$ as:
\begin{align*}
	D_1^{(V)} \mathcal{H}_{\sigma}(f, \kappa) = \int_{\mathbb{X}} |\nabla f|^{p-2} \nabla f \cdot \nabla V~d\mu 
																					+ \kappa \int_{\mathbb{X}} \partial_f \bar{L}_{\sigma}(f) \cdot V~d\mu.
\end{align*}
%
%
  
 By the Minimax Theorem, we have~$\mathcal{H}_{\sigma}(f^{\epsilon, p}, \kappa^{\epsilon, p})
 = \inf_{f} \sup_{\kappa} \mathcal{H}_{\sigma}(f, \kappa) 
 = \sup_{\kappa} \inf_{f} \mathcal{H}_{\sigma}(f, \kappa)$,
 where the infimum is taken over~$W^{1,p}((\mathbb{X}, \mu), \mathbb{Y})$
 and the supremum over~$\realnonnegative$.
 We therefore have~$\mathcal{H}_{\sigma}(f^{\epsilon, p}, \kappa^{\epsilon, p}) \geq \mathcal{H}_{\sigma}(f^{\epsilon, p}, 0)$,
 which yields the condition~$\kappa^{\epsilon, p} \left( L_{\sigma}(f^{\epsilon, p}) - (J^*_{\sigma}(\alpha) + \epsilon) \right) \geq 0$. 
 Moreover, from feasibility, we have~$L_{\sigma}(f^{\epsilon, p}) \leq J^*_{\sigma}(\alpha) + \epsilon$ 
 and~$\kappa^{\epsilon, p} \geq 0$, which implies 
 that~$\kappa^{\epsilon, p} \left( L_{\sigma}(f^{\epsilon, p}) - (J^*_{\sigma}(\alpha) + \epsilon) \right) \leq 0$.
 This results in the \textit{complementary slackness} 
 condition~$\kappa^{\epsilon, p} \left( L_{\sigma}(f^{\epsilon, p}) - (J^*_{\sigma}(\alpha) + \epsilon) \right) = 0$.
 From the Minimax equality, we get that~$(f^{\epsilon, p}, \kappa^{\epsilon, p})$ is also a 
 critical point of~$\mathcal{H}_{\sigma}$, that is $D_1^{(V)} \mathcal{H}_{\sigma}(f^{\epsilon, p}, \kappa^{\epsilon, p}) = 0$
 for any~$V \in W^{1,p}((\mathbb{X},\mu), \mathbb{Y})$:
\begin{align*}
	0 &= \int_{\mathbb{X}} |\nabla f^{\epsilon, p}|^{p-2} \nabla f^{\epsilon, p} \cdot \nabla V~d\mu 
																					+ \kappa^{\epsilon, p} \int_{\mathbb{X}} \partial_f \bar{L}_{\sigma}(f^{\epsilon, p}) \cdot V~d\mu \\
		&= - \int_{\mathbb{X}} \frac{1}{\mu} \nabla \cdot \left( \mu |\nabla f^{\epsilon, p}|^{p-2} \nabla f^{\epsilon, p} \right) \cdot V~d\mu  
									+  \int_{\partial \mathbb{X}} |\nabla f^{\epsilon, p}|^{p-2} \nabla f^{\epsilon, p} \cdot \mathbf{n} V \mu~dS \\
																		&\qquad \qquad			+ \kappa^{\epsilon, p} \int_{\mathbb{X}} \partial_f \bar{L}_{\sigma}(f^{\epsilon, p}) \cdot V~d\mu,
\end{align*}
where we have used the Divergence Theorem to obtain the final
equality, with~$S$ as the surface measure on~$\partial \mathbb{X}$. 
This is the \textit{stationarity} condition.
As the above holds for any variation~$V \in W^{1,p}((\mathbb{X}, \mu); \mathbb{Y)}$,
it must follow that~$-\frac{1}{\mu} \nabla \cdot \left( \mu |\nabla f^{\epsilon, p}|^{p-2} \nabla f^{\epsilon, p} \right) 
+ \kappa^{\epsilon, p} \partial_f \bar{L}_{\sigma}(f^{\epsilon, p}) = 0$ $\mu$-a.e. in~$\mathbb{X}$ 
  and~$\mu \nabla f^{\epsilon, p} \cdot \mathbf{n} = 0$ on~$\partial \mathbb{X}$,
and if we do not suppose stronger regularity of~$f^{\epsilon, p}$,
the equations must be hold weakly.  

The above correspond to the necessary KKT conditions. Conversely, any
solution pair $(f^{\epsilon, p}, \kappa^{\epsilon, p})$ which satisfies the above
KKT conditions is a saddle point for the Lagrangian~$\mathcal{H}_\sigma$ and is
a solution to the original optimization problem. 
%
%
\section{Proof of Theorem~\ref{thm:conv_W^1,p_Lip} (Convergence as~$p \rightarrow \infty$)}
%
%
\emph{(i) Monotonicity properties of~$W^{1,p}((\mathbb{X}, \mu); \mathbb{Y})$.}
We first note that for~$p,q \in \mathbb{N}$, $1 < p < q$
and an~$f \in W^{1, p} ((\mathbb{X},\mu); \mathbb{Y})$,
$\left \| | f| \right \|_{L^p(\mathbb{X},\mu)} 
\leq \left \| | f| \right \|_{L^q(\mathbb{X},\mu)}$ and
$\left \| | \nabla f| \right \|_{L^p(\mathbb{X},\mu)} 
\leq \left \| | \nabla f| \right \|_{L^q(\mathbb{X},\mu)}$.
It follows that~$W^{1, q} ((\mathbb{X},\mu); \mathbb{Y}) 
\subseteq W^{1, p} ((\mathbb{X},\mu); \mathbb{Y})$.
In particular, for any~$p \in \mathbb{N}$, $p > 1$,
we have~$\left \| | f| \right \|_{L^p(\mathbb{X},\mu)} 
\leq \left \| | f| \right \|_{L^\infty(\mathbb{X},\mu)}$,
$\left \| | \nabla f| \right \|_{L^p(\mathbb{X},\mu)} 
\leq \left \| | \nabla f| \right \|_{L^\infty(\mathbb{X},\mu)}$
and~$W^{1, \infty} ((\mathbb{X},\mu); \mathbb{Y}) 
\subseteq W^{1, p} ((\mathbb{X},\mu); \mathbb{Y})$.
It then follows that $\left \lbrace f \in W^{1, q} ((\mathbb{X},\mu); \mathbb{Y}) \; | \; L_{\sigma}(f)  \leq \epsilon \right \rbrace 
\subseteq \left \lbrace f \in W^{1, p} ((\mathbb{X},\mu); \mathbb{Y}) \; | \; L_{\sigma}(f)  \leq \epsilon \right \rbrace$
for~$1 < p < q \leq \infty$.

\emph{(ii) Minimizers.}
From the strict convexity of~$L_{\sigma}$,
it follows that $\left \lbrace f \in W^{1, p} ((\mathbb{X},\mu); \mathbb{Y}) \; | \; L_{\sigma}(f)  \leq \epsilon \right \rbrace$
is closed and convex for any~$1 < p \leq \infty$.
Moreover, the semi-norm of~$f \in W^{1,p}((\mathbb{X}, \mu); \mathbb{Y})$,
i.e., $\| |\nabla f| \|_{L^p(\mathbb{X}, \mu)}$,
is convex. The existence of global minimizers for the problem:
\begin{align*}
	\inf_{\substack{f \in W^{1,p}((\mathbb{X},\mu);\mathbb{Y})}}~ \left \lbrace \left \| | \nabla f| \right \|_{L^p(\mathbb{X},\mu)},
	\qquad \text{s.t.}~~ L_{\sigma}(f)  \leq J^*_{\sigma}(\alpha) + \epsilon \right \rbrace
\end{align*}
was established in Section~\ref{sec:saddle_point_proof_Sec_3} for
every~$\dim(\mathbb{X}) < p \leq \infty$ and~$\epsilon > 0$.

\emph{(iii) Monotonicity of minimum value.}
From the existence of a global minimum value
for any~$\dim(\mathbb{X}) < p < \infty$, and the monotonicity properties of
$W^{1,p}(\mathbb{X}, \mu)$, we get for~$\dim(\mathbb{X}) < p \leq q$:
\begin{align*}
	\min_{\substack{f \in W^{1,p}((\mathbb{X},\mu);\mathbb{Y}) \\ L_{\sigma}(f)  \leq J^*_{\sigma}(\alpha) + \epsilon}}~ \left \| | \nabla f| \right \|_{L^p(\mathbb{X},\mu)}
\; \leq \; \min_{\substack{f \in W^{1,q}((\mathbb{X},\mu);\mathbb{Y}) \\ L_{\sigma}(f)  \leq J^*_{\sigma}(\alpha) + \epsilon}}~ \left \| | \nabla f| \right \|_{L^q(\mathbb{X},\mu)}.
\end{align*}
In particular, we get for any~$p > \dim(\mathbb{X})$:
\begin{align*}
		\min_{\substack{f \in W^{1,p}((\mathbb{X},\mu);\mathbb{Y}) \\ L_{\sigma}(f)  \leq J^*_{\sigma}(\alpha) + \epsilon}}~ \left \| | \nabla f| \right \|_{L^p(\mathbb{X},\mu)}
\; \leq \; \min_{\substack{f \in \Lip((\mathbb{X},\mu);\mathbb{Y}) \\  L_{\sigma}(f)  \leq J^*_{\sigma}(\alpha) + \epsilon }} ~\lip \left( f \right).
\end{align*}
Therefore, by the convergence of bounded monotone sequences, we get:
\begin{align*}
	\lim_{p \rightarrow \infty} ~ \min_{\substack{f \in W^{1,p}((\mathbb{X},\mu);\mathbb{Y}) \\ L_{\sigma}(f)  \leq J^*_{\sigma}(\alpha) + \epsilon}}~ \left \| | \nabla f| \right \|_{L^p(\mathbb{X},\mu)}
\; \leq \; \min_{\substack{f \in \Lip((\mathbb{X},\mu);\mathbb{Y}) \\  L_{\sigma}(f)  \leq J^*_{\sigma}(\alpha) + \epsilon }} ~\lip \left( f \right) = \bar{\alpha}(\epsilon).
\end{align*}
%
%

\emph{(iv) Upper bound is indeed the supremum.}
We now consider the sequence of minimizers~$\left \lbrace f^{\epsilon,p}_{\sigma} \right \rbrace_{p \in \mathbb{N}}$:
\begin{align*}
	f^{\epsilon,p}_{\sigma} \in \arg \min_{\substack{f \in W^{1,p}((\mathbb{X},\mu);\mathbb{Y}) \\ L_{\sigma}(f)  \leq J^*_{\sigma}(\alpha) + \epsilon}}~ \left \| | \nabla f| \right \|_{L^p(\mathbb{X},\mu)}.
\end{align*}
Fixing a $p > \dim(\mathbb{X})$,
from the monotonicity of minimum values
and the compactness of~$\mathbb{Y}$, we get
that the sequence~$\left \lbrace f^{\epsilon,q}_{\sigma} \right \rbrace_{q \geq p}$
is uniformly bounded in~$W^{1,p}((\mathbb{X}, \mu),\mathbb{Y})$
as~$\| f^{\epsilon,q}_{\sigma} \|_{W^{1,p}((\mathbb{X}, \mu),\mathbb{Y})} \leq M + \bar{\alpha}(\epsilon)$.
Moreover, for~$\dim(\mathbb{X}) < p \leq \infty$, we
have from Morrey's inequality that:
\begin{align*}
	| f^{\epsilon,p}_{\sigma}(x_1) - f^{\epsilon,p}_{\sigma}(x_2) | 
&\leq \frac{2p\dim(\mathbb{X})}{p-\dim(\mathbb{X})}~ | x_1 - x_2 |^{1 - \frac{\dim(\mathbb{X})}{p}} \left \| | \nabla f^{\epsilon,p}_{\sigma} | \right \|_{L^p(\mathbb{X},\mu)} \\
&\leq 2 C \dim(\mathbb{X}) \left( 1+ \dim(\mathbb{X}) \right)~| x_1 - x_2 |^{\frac{1}{1+ \dim(\mathbb{X})}}~ \bar{\alpha}(\epsilon),
\end{align*}
where~$C = \max \left \lbrace 1 , \mathrm{diam}(\mathbb{X})^{\frac{\dim(\mathbb{X})}{1+ \dim(\mathbb{X})}} \right \rbrace$.
It follows from the above that the sequence~$\left \lbrace f^{\epsilon,p}_{\sigma} \right \rbrace_{p \in \mathbb{N}, p > \dim(\mathbb{X})}$
is also uniformly equicontinuous. Therefore, by the Arzel{\`a}-Ascoli Theorem~\cite{WR:64},
there exists a subsequence~$\left \lbrace f^{\epsilon,p_j}_{\sigma} \right \rbrace_{j \in \mathbb{N}}$
that converges uniformly to a Lipschitz
continuous~$f^{\epsilon,\infty}_{\sigma}$. 
Moreover, from the monotonicity of minimum values,
it follows that the Lipschitz constant~$ \lip( f^{\epsilon,\infty}_{\sigma}) = 
\| | \nabla f^{\epsilon,\infty}_{\sigma} | \|_{L^{\infty}(\mathbb{X}, \mu)} \leq \bar{\alpha} (\epsilon)$.
We also have~$\lip(f^{\epsilon,\infty}_{\sigma}) \geq 
\min_{\substack{f \in \Lip((\mathbb{X},\mu);\mathbb{Y}) \\  L_{\sigma}(f)  \leq J^*_{\sigma}(\alpha) + \epsilon }} ~\lip \left( f \right) = \bar{\alpha}(\epsilon)$.
Therefore, we have~$\lip(f^{\epsilon,\infty}_{\sigma}) = \bar{\alpha}(\epsilon)$,
and~$\left \lbrace f^{\epsilon,p}_{\sigma} \right \rbrace_{p \in \mathbb{N}}$
converges uniformly (upto a subsequence) to a (global) minimizer~$f^{\epsilon, \infty}$
of~\eqref{eq:lipschitz_min}.
%

%
%

\section{Numerical analysis of classifier robustness}
\begin{figure}[H]
  \centering
    \includegraphics[width=1\columnwidth]{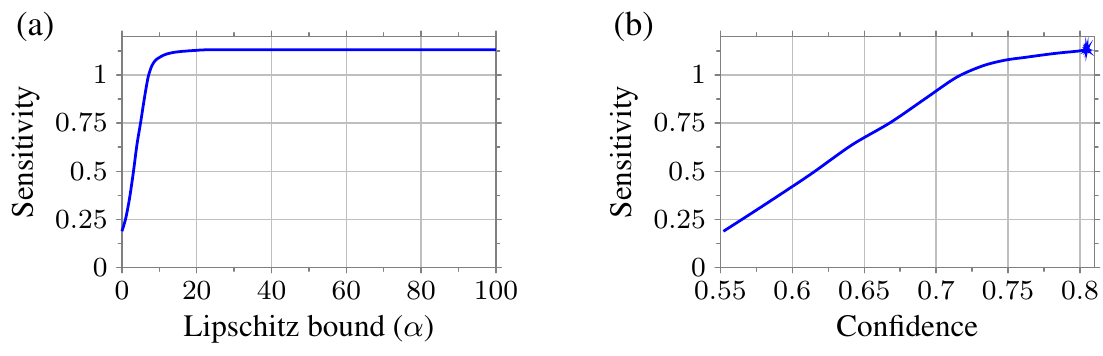} 
    \caption[]{For the classification problem discussed in Section
      \ref{sec: robust_min} and Fig. \ref{fig:feedbackInterconnection} in the main manuscript, (a) shows the classifier's sensitivity to data perturbation as a function of the Lipschitz bound, the plot shows that sensitivity increases with the Lipschitz bound up to a certain value ($\alpha=18$). (b) shows the tradeoff between performance and robustness, seen by the monotonic increase of the sensitivity as a function of classification confidence.}
  \label{fig:sens_vs_lips_conf} 
\end{figure}

In this section, we provide numerical analysis to quantify a classifier's robustness against data perturbation for the classification problem discussed in Section \ref{sec: robust_min} and Fig. \ref{fig:feedbackInterconnection} of the manuscript. Using the same setup explained in Section \ref{sec: robust_min}, we design our classifiers by constructing a graph $\mathcal{G}=(\mathcal{V},\mathcal{E})$ with $n=500$ randomly selected nodes by connecting each node to its $10$ nearest neighbors. We compute the solution $\mathbf{v}^*$ to
\eqref{eq:p-d_Lipschitz_const_loss_min} for different values of the
Lipschitz constant $\alpha \in (0,100]$. We generate a nominal testing set of $1000$
i.i.d. samples from $\sigma$, associate them with the closest node,
and evaluate the nominal classification confidence of $\mathbf{v}^*$. Then, we perturb each testing data sample with $\delta \in \mathbb{R}^2$ with $\| \delta \|_2 = 0.05$ in the direction perpendicular to the closest edge, associate each perturbed data point with the closest node and evaluate the perturbed classification confidence. To measure the sensitivity of the designed classifier, we compute the norm of the difference between the nominal and the perturbed confidence, then appoint the sensitivity measure to the maximum value across all the testing data points. Fig. \ref{fig:sens_vs_lips_conf}(a) shows the plot of the sensitivity for each classifier designed using different Lipschitz bound $\alpha$, it can be seen that the sensitivity increases as we increase the Lipschitz bound up to $\alpha=18$. Fig. \ref{fig:sens_vs_lips_conf}(b) shows the plot of the sensitivity for each classifier as a function of the classification confidence, we observe a tradeoff between classification performance and robustness to data perturbation seen by the monotonic increase of the sensitivity as a function of classification confidence, where improving classification performance comes at the expenses of robustness to data perturbation.
      
\end{appendices}


\end{document}